\documentclass{article}

\usepackage{arxiv}

\usepackage{hyperref}
\usepackage{url}

\usepackage{Sty/mcr}
\usepackage{bm}
\usepackage{amsmath}
\usepackage{amssymb}
\usepackage{mathtools}
\usepackage{amsthm}
\mathtoolsset{showonlyrefs}
\usepackage{subcaption}
\usepackage{cite}
\usepackage{booktabs}
\usepackage{longtable}

\usepackage[utf8]{inputenc} 
\usepackage[T1]{fontenc}    
\usepackage{hyperref}       
\usepackage{url}            
\usepackage{booktabs}       
\usepackage{amsfonts}       
\usepackage{nicefrac}       
\usepackage{microtype}      
\usepackage{cleveref}       
\usepackage{lipsum}         
\usepackage{graphicx}
\usepackage{natbib}
\usepackage{doi}

\usepackage{placeins}

\theoremstyle{plain}
\newtheorem{theorem}{Theorem}[section]

\theoremstyle{definition}

\theoremstyle{remark}

\title{Statistical Testing for Multiple Instance Learning via Selective Inference with Applications to Computational Pathology}

\usepackage{authblk}

\author[1]{%
	Noriaki Hashimoto\thanks{\texttt{noriaki.hashimoto.jv@riken.jp}}%
}
\author[2,1]{%
	Shuichi Nishino
}
\author[2]{%
	Teruyuki Katsuoka
}
\author[2,1]{%
	Tomohiro Shiraishi
}
\author[2]{%
	Daiki Miwa
}
\author[2]{%
	Hiroyuki Hanada
}
\author[3,1]{%
	Jun Sakuma
}
\author[4]{%
	Hidekata Hontani
}
\author[5]{%
	Hiroaki Miyoshi
}
\author[2,1]{%
	Ichiro Takeuchi\thanks{\texttt{takeuchi.ichiro.n6@f.mail.nagoya-u.ac.jp}}%
}
\affil[1]{RIKEN}
\affil[2]{Nagoya University}
\affil[3]{Institute of Science Tokyo}
\affil[4]{Nagoya Institute of Technology}
\affil[5]{Kurume University}

\begin{document}

\maketitle

\begin{abstract}

Multiple instance learning (MIL) is widely used in computational
pathology because it enables weakly supervised analysis of whole-slide
images (WSIs) without requiring patch-level annotations.
In attention-based MIL, instances with high attention scores are often
interpreted as diagnostically important regions and used as visual
explanations.
However, attention scores alone cannot determine whether selected high-attention instances are significantly different from normal instances, limiting the reliability of attention-based explanations.
In this paper, we formulate the evaluation of high-attention instances
as a statistical hypothesis testing problem.
Specifically, we assess whether a selected high-attention instance significantly deviates from a representative normal reference instance selected based on feature similarity.
A major challenge is that both the target instance and the reference
instance are selected through data-dependent procedures, rendering
standard hypothesis testing invalid.
To address this issue, we introduce a selective inference (SI) framework that explicitly accounts for the selection events induced by attention-based instance selection and adaptive reference selection, thereby enabling the computation of valid selective $p$-values conditional on these events.
Experiments demonstrate Type-I error control on synthetic and
MNIST-based data and practical applicability to pathological WSIs,
with higher statistical power than the conventional
over-conditioning approach.
\end{abstract}


\section{Introduction}

Multiple instance learning (MIL)~\citep{dietterich1997solving,maron1997framework} is a weakly supervised learning framework in which labels are assigned to sets of instances, referred to as \emph{bags}, while the individual instances within each bag remain unlabeled.
MIL has become one of the standard
approaches for analyzing whole-slide images (WSIs) in computational
pathology~\citep{das2018multiple,yamamoto2019automated,campanella2019clinical,hashimoto2020multi}.
Because WSIs are extremely large, they are typically divided into a
large number of image patches for computational analysis.
While obtaining dense patch-level annotations for these patches from
pathologists is prohibitively expensive, MIL allows the image patches
to be treated as instances and slide-level diagnoses, such as tumor or
normal, to be used as bag labels.
This formulation enables the development of accurate diagnostic models
without requiring exhaustive manual annotation.

Among various MIL approaches~\citep{wu2015deep,li2018thoracic,campanella2019clinical}, attention-based MIL (ABMIL)~\citep{ilse2018attention} has become one of the most widely used architectures due to its simplicity, interpretability, and competitive empirical performance.
Although more advanced MIL architectures have recently been proposed, including transformer-based and graph-based models, ABMIL remains one of the most widely studied and interpretable MIL frameworks.
ABMIL assigns an attention score to each instance and aggregates instance features according to the learned attention weights.
Consequently, instances with high attention scores are often interpreted as diagnostically important regions and are widely used as visual explanations in computational pathology~\citep{sudharshan2019multiple,campanella2019clinical,hashimoto2020multi,li2021dual,
lu2021data,zhang2022dtfd,takagi2023transformer,hashimoto2024multimodal}.
Attention maps are increasingly utilized not only for model interpretation but also for supporting downstream clinical analysis and identifying candidate abnormal regions.

Despite their widespread use, the reliability of attention-based explanations remains an open question.
Recent studies have argued that attention scores do not necessarily correspond to causal or clinically meaningful evidence~\citep{jain2019attention,serrano2019attention,wiegreffe2019attention}.
In computational pathology, high-attention patches may arise due to dataset-specific biases, staining artifacts, or optimization effects rather than genuine pathological abnormalities~\citep{wang2022label,cai2024rethinking,sun2025label,zehnder2025diagnostic}.
Therefore, while attention scores provide a useful ranking of candidate regions, additional statistical evidence is necessary to assess whether a selected high-attention instance differs significantly from normal instances.

To address this issue, we formulate the evaluation of high-attention instances as a statistical hypothesis testing problem.
Because normal instances can exhibit substantial heterogeneity, comparing a target instance with a fixed or globally representative normal instance may not provide an appropriate reference.
Instead, for each target instance, we adaptively retrieve similar normal instances and select a representative instance among them as a local reference.
We then assess whether the selected high-attention instance significantly differs from this adaptively selected normal reference using a distance-based statistical test.
The adaptive nature of this selection procedure gives rise to the statistical challenges considered in this paper.

A major challenge is that both the target instance and the reference instance are selected adaptively from the observed data.
The target instance is selected based on its attention score, while a representative normal instance is adaptively selected from normal instances similar to the target instance.
As a result, the statistical test is conducted only after a sequence of data-dependent selection procedures.
Since standard hypothesis tests assume that the hypothesis to be tested
is specified independently of the data used for testing, naively
applying them while ignoring the data-dependent selection of the target
and reference instances leads to invalid statistical inference~\citep{kriegeskorte2009circular}.
Therefore, properly accounting for the adaptive selection process is essential for obtaining statistically valid significance measures.

\begin{figure}[tb]
\begin{center}
      \includegraphics[width=0.85\linewidth]{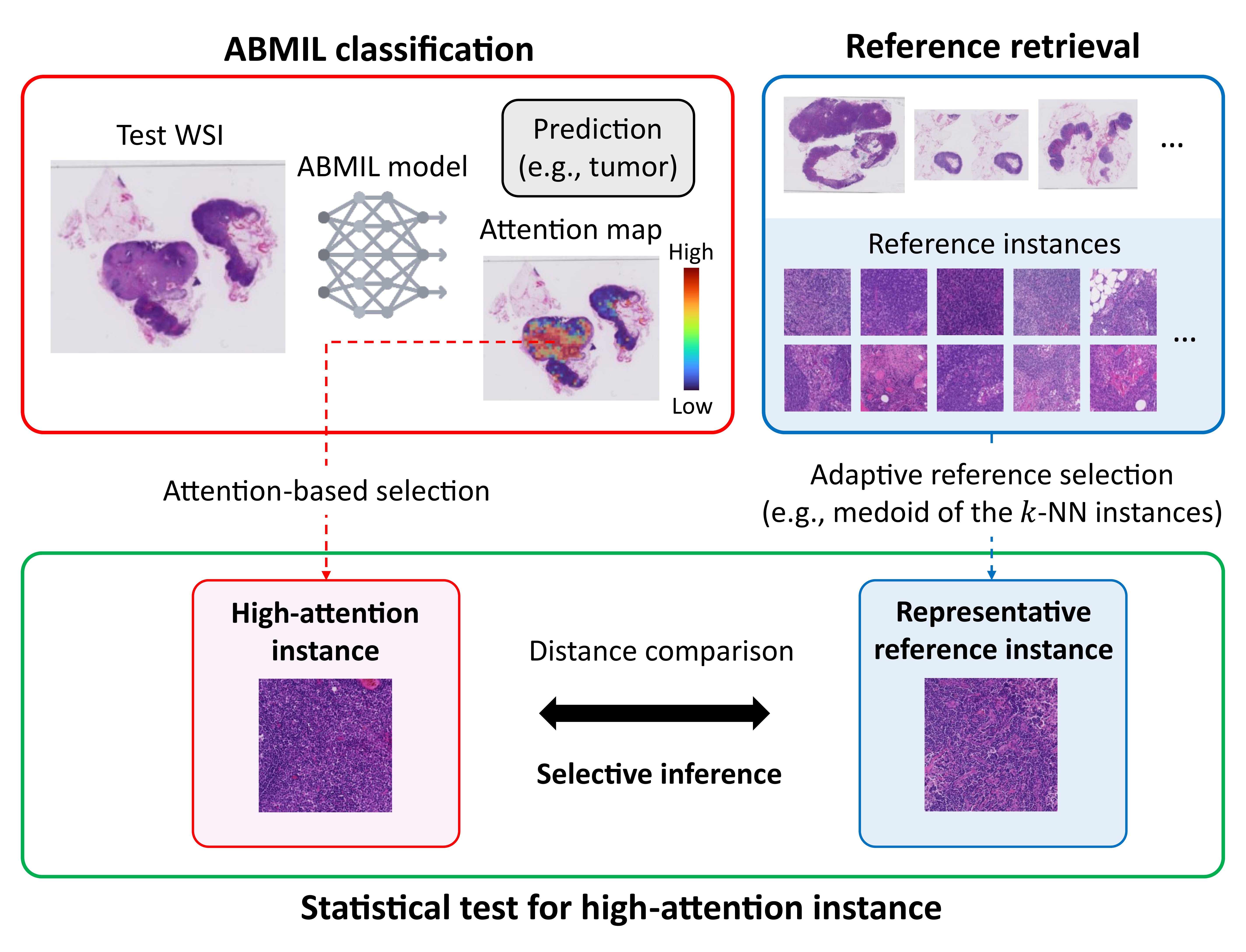}
\end{center}
\caption{
Overview of the proposed framework.
An ABMIL model predicts the bag label and generates an attention map for a test WSI.
A high-attention instance is selected according to its attention score, and a representative reference instance is adaptively selected from the reference instances.
The proposed SI framework then evaluates whether the selected high-attention instance significantly differs from the representative reference instance while accounting for the adaptive selection procedures.
}
\label{fig:GA}
\end{figure}

To overcome this challenge, we introduce selective inference (SI)~\citep{lee2016exact,tian2018selective,taylor2018post}, a statistical framework for valid inference after data-dependent selection.
SI performs hypothesis testing conditional on the selection event itself and has recently been extended to a variety of machine learning problems, including various types of deep neural networks~\citep{duy2022quantifying,miwa2023salient,shiraishi2024statistical,shiraishi2026statistical,niihori2025quantifying}.
In this work, we explicitly characterize the selection events induced by the data-dependent selection of the target instance and its normal reference, and derive valid conditional $p$-values for evaluating high-attention instances.
Figure~\ref{fig:GA} illustrates the overall workflow of the proposed framework.

We validate the proposed framework through synthetic experiments, MNIST-based proof-of-concept experiments, and real-world pathological WSI analysis.
Experiments demonstrate Type-I error control on synthetic and
MNIST-based data and practical applicability to pathological WSIs,
with higher statistical power than the conventional
over-conditioning approach.

To the best of our knowledge, this study is the first attempt to quantify the statistical significance of instances selected by MIL models within a statistical hypothesis testing framework.
The key idea is to regard the data-dependent instance selection induced by an MIL algorithm as a selection event and to account for this event within the SI framework.
This principle is not specific to ABMIL and can, in principle, be extended to other MIL algorithms for which the corresponding selection events can be characterized.
In this study, we focus on ABMIL as a proof-of-concept for statistically evaluating MIL-selected instances using SI.
We further use WSI-based pathological image analysis, which motivated this study, as a working example to demonstrate the practical relevance of the proposed framework.

\paragraph{Related work.}

MIL was originally introduced as a framework for learning from labeled bags of unlabeled instances~\citep{dietterich1997solving,maron1997framework,zhou2002neural,andrews2002support} and has since become a standard paradigm in computational pathology.
Existing MIL approaches can broadly be categorized into instance-based methods and embedding-based methods.
Early approaches often relied on max pooling or average pooling to aggregate instance-level information~\citep{wu2015deep,feng2017deep,maron1997framework,
li2018thoracic,wang2018revisiting,cruz2014automatic,shrivastava2015generalized,
kraus2016classifying}, while more recent methods employ learnable aggregation mechanisms.

Among these approaches, ABMIL~\citep{ilse2018attention} has become particularly influential due to its ability to assign adaptive importance weights to instances.
Numerous studies have applied ABMIL to computational pathology, where high-attention patches are interpreted as tumor regions or diagnostically important tissues~\citep{hashimoto2020multi,li2021dual,zhang2022dtfd}.
More recently, transformer-based MIL architectures~\citep{lee2019set,chen2021multimodal,
shao2021transmil,chen2022scaling,takagi2023transformer} have further improved slide-level predictive performance by modeling interactions among instances using self-attention mechanisms~\citep{vaswani2017attention}.
However, the goal of the present study is not to develop a new MIL architecture, but to establish a statistically principled framework for evaluating instances deemed important by MIL models.
We focus on ABMIL as a proof-of-concept because its explicit attention mechanism provides a natural instance-selection mechanism.
For this purpose, ABMIL provides a suitable model because its attention mechanism is explicit, widely adopted, and directly used for instance-level interpretation.

Despite these advances, several studies have questioned whether attention scores faithfully reflect meaningful instance-level importance~\citep{jain2019attention,serrano2019attention,wiegreffe2019attention}.
In particular, attention maps may reflect optimization artifacts or dataset-specific biases rather than clinically meaningful evidence.
These observations motivate the need for statistically principled methods to evaluate the reliability of selected high-attention instances.

SI provides a principled framework for valid statistical inference after data-dependent selection by conditioning inference on the selection event induced by the underlying algorithm~\citep{taylor2015statistical,lee2016exact,tibshirani2016exact}.
Much of the early SI literature focused on model and feature selection in regression, particularly sparse models such as the Lasso, but the framework has subsequently been extended to increasingly complex feature-selection and model-selection procedures~\citep{terada2023selective,hyun2018exact,panigrahi2024selective,suzumura2017selective,yamada2018post,charkhi2018asymptotic,shiraishi2025statistical}.
Beyond supervised feature selection, SI has also been developed for a broad range of data-dependent inference problems in which the objects or hypotheses to be tested are themselves identified from the observed data, including clustering, changepoint detection, outlier and anomaly detection, and principal component analysis~\citep{chen2023selective,gao2024selective,duy2020computing,chen2020valid,perry2026inference}.
In parallel, methodological developments such as randomized SI and parametric-programming-based approaches have been proposed to reduce unnecessary conditioning and improve statistical power while retaining valid SI~\citep{tian2018selective,le2022more}.
These developments have substantially broadened the range of data-dependent algorithms for which valid post-selection inference can be performed.

More recently, SI has been extended to substantially more complex selection mechanisms involving deep neural networks.
Representative applications include statistical inference for neural-network-based image segmentation and salient-region detection~\citep{duy2022quantifying,miwa2023salient}, deep nearest-neighbor anomaly detection~\citep{niihori2025quantifying}, saliency analysis for graph neural networks~\citep{nishino2025statistical}, and attention-based hypotheses generated by transformer models~\citep{shiraishi2024statistical,shiraishi2026statistical}.
Computational frameworks that automate the characterization of complex neural-network-induced selection events have also been developed, further expanding the applicability of SI to modern deep-learning models~\citep{katsuoka2025si4onnx}.
These advances motivate the use of SI for MIL, where instances regarded as important by an MIL algorithm are themselves selected in a data-dependent manner.
In our setting, both the target instance and its normal reference are adaptively selected from the observed data, and directly testing them without accounting for these selection procedures leads to selection bias.
We therefore incorporate the corresponding selection events into the SI framework to obtain valid selective $p$-values.
To the best of our knowledge, this is the first study to formulate statistical hypothesis testing for the significance of instances identified as important by MIL algorithms within an SI framework.

\paragraph{Contributions.}

The main contributions of this paper are summarized as follows:

\begin{itemize}

\item We formulate the evaluation of high-attention instances in ABMIL as a statistical hypothesis testing problem, enabling quantitative assessment of instance-level significance.

\item We develop an SI framework that accounts for both attention-based instance selection and adaptive reference selection, enabling valid post-selection inference for high-attention instances.

\item Experiments demonstrate Type-I error control on synthetic and
MNIST-based data and practical applicability to pathological WSIs,
with higher statistical power than the conventional
over-conditioning approach.

\end{itemize}

\section{Problem setup: multiple instance learning}
\label{sec:mil}

\subsection{MIL-based classification problem}
\label{sec:mil_problem}

MIL is a weakly supervised learning framework in which class labels are assigned only to \emph{bags}, each consisting of a set of \emph{instances}.
Let $N$ denote the number of bags in a dataset, and define $[N] := \{1, \ldots, N\}$ as the index set of the bags.
For each $n \in [N]$, let $M^n$ denote the number of instances in the $n$-th bag, and define $[M^n] := \{1, \ldots, M^n\}$ as the corresponding instance index set.
We denote the $n$-th bag by
\begin{align*}
X^n = \{\bm{x}^{n,m}\}_{m \in [M^n]},
\end{align*}
where $\bm{x}^{n,m}$ denotes the $m$-th instance in the $n$-th bag.
Let $Y^n$ denote the class label assigned to bag $X^n$, and let $y^{n,m}$ denote the unobserved class label of instance $\bm{x}^{n,m}$.
In the standard binary MIL setting, a positive bag ($Y^n = 1$) contains at least one positive instance ($y^{n,m} = 1$), whereas a negative bag ($Y^n = 0$) contains only negative instances ($y^{n,m} = 0$).
Finally, let $f$ denote the MIL classification model, whose prediction for the $n$-th bag is given by $\hat{Y}^n = f(X^n)$.
The model is trained so that $\hat{Y}^n$ accurately predicts the corresponding bag label $Y^n$.

In computational pathology, the WSI of patient $n$ is represented as the \emph{bag} $X^n$, and each image patch tiled from the WSI is represented as an \emph{instance} $\bm{x}^{n,m}$.
We consider a binary classification problem distinguishing between cancerous and normal cases in pathological images.
A WSI from a cancer case typically contains both cancerous and normal image patches, whereas a WSI from a normal case contains only normal patches.
Accordingly, WSI classification can be naturally formulated as an MIL problem, where cancerous patches are regarded as positive instances and normal patches as negative instances.

\subsection{ABMIL and attention-based instance selection}
\label{sec:abmil}

\begin{figure}[tb]
\begin{center}
      \includegraphics[width=1.0\linewidth]{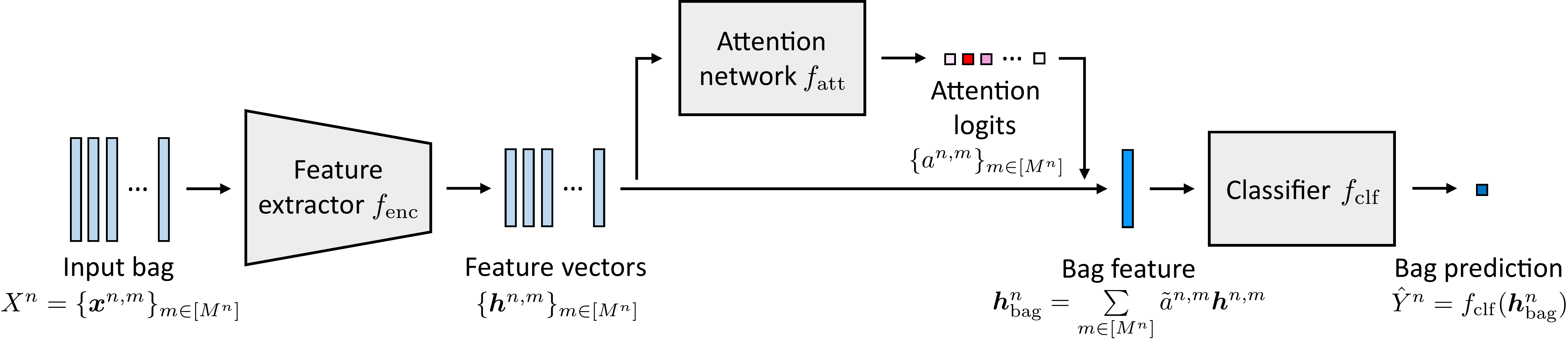}
\end{center}
\caption{
Overview of the ABMIL model.
An input bag $X^n=\{\bm x^{n,m}\}_{m\in[M^n]}$ is first mapped to instance
feature representations $\{\bm h^{n,m}\}_{m\in[M^n]}$ by the feature extractor
$f_{\mathrm{enc}}$.
The attention network $f_{\mathrm{att}}$ assigns an attention logit to
each instance, and the bag feature is computed as the attention-weighted
sum of the instance features.
The classifier $f_{\mathrm{clf}}$ predicts the bag label from the
aggregated bag feature.
Instances whose attention logits exceed the attention threshold are identified as high-attention instances and are subsequently subjected to selective inference.
}
\label{fig:MIL}
\end{figure}

In this study, we employ ABMIL~\citep{ilse2018attention} as the MIL classification model.
Figure~\ref{fig:MIL} illustrates the architecture of ABMIL.
ABMIL consists of three primary components: a feature extractor $f_{\mathrm{enc}}$, an attention network $f_{\mathrm{att}}$, and a classifier $f_{\mathrm{clf}}$.
The model first extracts instance-level feature representations, computes attention logits for individual instances, and then aggregates the instance features into a bag-level representation for classification.

The feature extractor $f_{\mathrm{enc}}$ maps an input image
$\bm{x}^{n,m}\in\mathbb R^{d}$
to a $d'$-dimensional feature representation:
\begin{align}
\bm{h}^{n,m} = f_{\mathrm{enc}}(\bm{x}^{n,m}).
\end{align}

The attention network $f_{\mathrm{att}}$ computes an attention logit
\begin{align}
a^{n,m}=f_{\mathrm{att}}(\bm{h}^{n,m}),
\end{align}
for each instance.
During training, the attention logits are transformed by the
sigmoid function:
\begin{align}
a'^{\,n,m}
=
\operatorname{sigmoid}(a^{n,m}),
\end{align}
where $a'^{\,n,m}\in(0,1)$ denotes the post-sigmoid attention score.
The normalized attention weights are then computed by applying the
softmax function to the post-sigmoid attention scores:
\begin{align}
\tilde a^{n,m}
=
\frac{\exp(a'^{\,n,m})}
{\sum_{m'=1}^{M^n}\exp(a'^{\,n,m'})}.
\end{align}
%

The instance features $\{\bm{h}^{n,m}\}_{m \in [M^n]}$ are aggregated into a bag-level representation by attention-weighted pooling:
\begin{align}
\bm{h}_{\mathrm{bag}}^n
=
\sum_{m \in [M^n]}
\tilde a^{n,m} \bm{h}^{n,m}.
\end{align}

Using the classifier $f_{\mathrm{clf}}$, the final prediction is obtained as
\begin{align}
\hat{Y}^n
=
f(X^n)
=
f_{\mathrm{clf}}(\bm{h}_{\mathrm{bag}}^n).
\end{align}

The entire ABMIL model is trained end-to-end to minimize the classification error between the predicted labels $\hat{Y}^n$ and the corresponding bag labels $Y^n$.

In computational pathology, high-attention instances, corresponding to
image patches within a WSI, are often interpreted as diagnostically
important regions, such as areas containing tumor tissue~\citep{hashimoto2020multi,li2021dual,lu2021data}.
Consequently, MIL attention maps are widely used as visual explanations for WSI classification models.
However, recent studies have questioned whether attention scores faithfully represent clinically meaningful evidence~\citep{jain2019attention,serrano2019attention,wiegreffe2019attention}.
Since the attention network is optimized solely for bag-level classification, high attention scores do not provide statistical evidence that the corresponding instances differ from normal instances.
In practice, high attention scores may also be influenced by dataset-specific biases, staining artifacts, or optimization effects, making it difficult to assess the statistical significance of selected instances based on attention scores alone.

In this study, we consider a test bag $X$ and focus on instances that receive high attention logits.
Let $\tau$ denote a fixed attention threshold.
Depending on the experimental setting, $\tau$ is either specified
a priori (e.g., $\tau=0$) or estimated beforehand from an independent
dataset.
We define
\begin{align}
\mathcal C(X)
=
\left\{
m \in [M]
\mid
a^m > \tau
\right\},
\end{align}
as the set of indices corresponding to high-attention instances in the bag $X$.
These instances are frequently used to explain the predictions of MIL models and are therefore natural targets for statistical investigation.
Our objective is to quantify the reliability of attention-based explanations by testing whether selected high-attention instances significantly deviate from a representative reference instance.
However, because both the target instance and the reference instance are selected adaptively from the data, standard hypothesis testing becomes invalid due to selection bias.
In the following sections, we introduce an SI framework for statistically evaluating high-attention instances while properly accounting for the adaptive selection process.

\section{Statistical testing formulation for high-attention instances}
\label{sec:test}

In this section, we formulate a statistical hypothesis testing problem for evaluating high-attention instances selected by a trained ABMIL model.
Let
\begin{align*}
X_{\mathrm {test}}
= 
\{\bm x_{\mathrm {test}}^i\}_{i \in [M_{\mathrm {test}}]}
\end{align*}
be a test bag and let
\begin{align*}
X_{\mathrm {ref}}
=
\{\bm x_{{\mathrm {ref}}}^j\}_{j \in [M_{\mathrm {ref}}]}
\end{align*}
be a reference set consisting of normal instances.
The reference set is assumed to be independent of the training data used for model fitting and the test bag used for statistical evaluation.
For each test instance, the trained ABMIL model computes a feature representation
$\bm h_{{\mathrm {test}}}^i = f_{\mathrm {enc}}(\bm x_{{\mathrm {test}}}^i)$
and an attention logit
$a_{{\mathrm {test}}}^i = f_{\mathrm {att}}(\bm h_{{\mathrm {test}}}^i)$.

As introduced in Section~\ref{sec:mil}, we focus on instances whose
attention logits are greater than the attention threshold $\tau$.
The set of high-attention instances in the test bag is defined as
\begin{align}
\label{eq:attention_set}
\mathcal C(X_{\mathrm {test}})
=
\{
i\in[M_{\mathrm {test}}]
\mid
a_{\mathrm {test}}^i
>
\tau
\}.
\end{align}
For each selected test instance $\bm x_{\mathrm {test}}^i$ with $i \in \mathcal C(X_{\mathrm {test}})$, we test whether it significantly deviates from a representative reference instance.

Normal tissue patches can exhibit substantial heterogeneity due to variations in tissue appearance, staining, slide preparation, and other dataset-specific factors.
Therefore, comparing a test instance with an arbitrary or globally representative normal instance may not provide an appropriate reference.
Instead, it is natural to select a representative normal instance from those that are similar to the test instance, thereby providing a local reference adapted to each test instance.
Several strategies can be considered for such adaptive reference selection.
In this study, we employ a $k$-nearest-neighbor-based medoid selection strategy to obtain the representative normal reference instance.

To define this procedure, let
$\bm h_{\mathrm {ref}}^j = f_{\mathrm {enc}}(\bm x_{\mathrm {ref}}^j)$
denote the feature representation of the $j$-th reference instance.
For each selected test instance, we identify its $k$ nearest normal instances in the feature space and select their medoid as the representative reference instance.
Let $j_{\mathrm {med}}\in[M_{\mathrm {ref}}]$ denote the index of the resulting medoid, and let $\bm x_{\mathrm {ref}}^{j_{\mathrm {med}}}$ denote the corresponding representative normal reference instance.

To formulate the testing problem, we regard each input instance as a realization of a random vector.
For a selected high-attention test instance
$\bm x_{\mathrm{test}}^i$
with
$i\in\mathcal C(X_{\mathrm{test}})$,
we omit the index $i$ hereafter for notational simplicity and denote the
selected instance by
$\bm x_{\mathrm{test}}$.
For the test instance
$\bm x_{\mathrm{test}}$,
we assume
\begin{align}
\bm X_{\mathrm{test}}
=
\bm s_{\mathrm{test}}
+
\bm\epsilon_{\mathrm{test}},
\end{align}
where
$\bm X_{\mathrm{test}}\in\mathbb R^d$
is the random vector corresponding to the observed instance
$\bm x_{\mathrm{test}}$,
$\bm s_{\mathrm{test}}\in\mathbb R^d$
is an unknown signal vector, and
$\bm\epsilon_{\mathrm{test}}
\sim
\mathcal N(\bm0,\sigma^2I_d)$
is an isotropic Gaussian noise vector.
Similarly, for each reference instance, we assume
\begin{align}
\label{eq:statistical_model_ref}
\bm X_{\mathrm {ref}}^j
=
\bm s_{\mathrm {ref}}^j
+
\bm\epsilon_{\mathrm {ref}}^j,
\end{align}
where $\bm s_{\mathrm{ref}}^j\in\mathbb R^d$ is an unknown signal vector and
$\bm\epsilon_{\mathrm{ref}}^j\sim\mathcal N(\bm 0,\sigma^2I_d)$
is independent of $\bm \epsilon_{\mathrm {test}}$.
The noise vectors $\bm\epsilon_{\mathrm{test}}$ and
$\{\bm\epsilon_{\mathrm{ref}}^j\}_{j\in[M_{\mathrm{ref}}]}$
are mutually independent.
The variance $\sigma^2$ is assumed known for the theoretical analysis.

It is important to note that this formulation does not assume that the underlying signal vectors themselves follow a Gaussian distribution or any other specific probability distribution.
The signal vectors $\bm s_{\mathrm{test}}$ and $\{\bm s_{\mathrm{ref}}^j\}_{j\in[M_{\mathrm{ref}}]}$ are treated as arbitrary unknown fixed vectors, and the Gaussian assumption is imposed only on the random perturbations around these signals.
Thus, the statistical model should be interpreted as a signal-plus-noise formulation, in which the hypothesis test concerns differences between the underlying signal vectors rather than differences between parameters of an assumed data-generating distribution.

We test whether the selected high-attention instance significantly deviates from the selected representative reference instance.
The null and alternative hypotheses are defined as
\begin{align}
\label{eq:hypothesis}
{\mathrm {H}}_0:
\bm s_{\mathrm {test}}
=
\bm s_{\mathrm {ref}}^{j_{\mathrm {med}}}
\quad
{\mathrm {vs.}}
\quad
{\mathrm {H}}_1:
\bm s_{\mathrm {test}}
\neq
\bm s_{\mathrm {ref}}^{j_{\mathrm {med}}}.
\end{align}

As the test statistic, we use the Euclidean distance between the selected test instance and the selected reference instance:
\begin{align}
\label{eq:test_statistic}
T\left(\bm X_{\mathrm {test}},\{\bm X_{\mathrm {ref}}^j\}_{j \in [M_{\mathrm {ref}}]}\right)
=
\frac{1}{
\sqrt2
}
\left\|
\bm X_{\mathrm {test}}
-
\bm X_{\mathrm {ref}}^{j_{\mathrm {med}}}
\right\|_2.
\end{align}
To remove the dependence on the noise scale $\sigma$ and obtain a test statistic whose null distribution does not depend on $\sigma$, we further define the standardized test statistic as
\begin{align}
\label{eq:standardized_test_statistic}
\widetilde T
\left(\bm X_{\mathrm {test}},\{\bm X_{\mathrm {ref}}^j\}_{j \in [M_{\mathrm {ref}}]}\right)
=
\frac{1}{
\sigma
}
T\left(\bm X_{\mathrm {test}},\{\bm X_{\mathrm {ref}}^j\}_{j \in [M_{\mathrm {ref}}]}\right)
.
\end{align}
The test statistic is written as a function of the selected test instance and the entire reference set because the identity of the representative reference instance is determined adaptively from the observed data.
To clarify why a conventional statistical test is invalid in this setting, first consider a hypothetical setting in which the target and reference instances are fixed in advance, independently of the observed data.
For any fixed reference index $j$, under the null hypothesis $\bm s_{\mathrm{test}} = \bm s_{\mathrm{ref}}^j$, we have
\begin{align*}
\bm X_{\mathrm{test}} - \bm X_{\mathrm{ref}}^j \sim \mathcal N(\bm 0, 2\sigma^2 I_d).
\end{align*}
Consequently,
\begin{align*}
\frac{1}{\sqrt{2}\sigma} \left\| \bm X_{\mathrm{test}} - \bm X_{\mathrm{ref}}^j \right\|_2 \sim \chi_d.
\end{align*}

A naive analysis would apply this fixed-reference null distribution to the adaptively selected target and reference instances as if their indices had been prespecified.
Let
\begin{align*}
\widetilde T_{\mathrm{obs}} = \widetilde T \left( \bm x_{\mathrm{test}}, \{\bm x_{\mathrm{ref}}^j\}_{j\in[M_{\mathrm{ref}}]} \right)
\end{align*}
denote the observed standardized test statistic after the data-dependent selection.
The resulting naive $p$-value is defined as
\begin{align}
\label{eq:naive_p}
p_{\mathrm{naive}} = \mathbb P \left( \chi_d \ge \widetilde T_{\mathrm{obs}} \right),
\end{align}
where $\chi_d$ denotes a chi-distributed random variable with $d$ degrees of freedom.
This $p$-value would be valid if the tested target and reference instances had been fixed independently of the data.
In the present setting, however, both instances are selected using the observed data, so the fixed-reference null distribution does not represent the post-selection distribution of the test statistic.
Consequently, $p_{\mathrm{naive}}$ does not in general control the Type-I error rate at the nominal level.

In the next section, we characterize the selection events induced by
attention-based target selection and adaptive reference selection,
and derive an SI framework for computing valid conditional $p$-values.

\section{Statistical inference framework}

In this section, we introduce an SI framework for the statistical test proposed in the previous section.
Our goal is to quantify the statistical significance of the high-attention instance while properly accounting for the data-dependent selection process.
Unlike the naive statistical test that ignores the selection procedure, the proposed framework computes selective $p$-values by conditioning on the selection event, thereby providing statistically valid inference.

\subsection{Vector-form test statistic}

For notational simplicity, we concatenate the input vectors of the test instance $\bm x_{\mathrm {test}}$ and all reference instances $\{\bm x_{\mathrm {ref}}^j\}_{j \in [M_{\mathrm {ref}}]}$ into a single vector
\begin{equation}
\bm y
=
\operatorname{vec}
(
\bm x_{\mathrm {test}},
\bm x_{\mathrm {ref}}^1,
\ldots,
\bm x_{\mathrm {ref}}^{M_{\mathrm {ref}}}
)
\in
\mathbb R^{(M_{\mathrm {ref}}+1)d},
\end{equation}
where $d$ denotes the input dimension and $M_{\mathrm {ref}}$ is the number of reference instances.
Similarly, let
\begin{equation}
\bm Y
=
\operatorname{vec}
(
\bm X_{\mathrm {test}},
\bm X_{\mathrm {ref}}^1,
\ldots,
\bm X_{\mathrm {ref}}^{M_{\mathrm {ref}}}
)
\in
\mathbb R^{(M_{\mathrm {ref}}+1)d},
\end{equation}
be the corresponding random vector.

Suppose that the selected representative reference instance is indexed
by $j_{\mathrm{med}}$.
The test statistic introduced in
Section~\ref{sec:test}
can then be written as
\begin{equation}
T(\bm Y)
=
\left\|
P\bm Y
\right\|_2,
\end{equation}
where $P\in\mathbb R^{(M_{\mathrm {ref}}+1)d\times (M_{\mathrm {ref}}+1)d}$ is a symmetric matrix defined as
\begin{equation}
P
=
\frac{1}{2}
\begin{pmatrix}
I_d & 0 & \cdots & -I_d & \cdots & 0\\
0 & 0 & \cdots & 0 & \cdots & 0\\
\vdots & & & & & \vdots\\
-I_d & 0 & \cdots & I_d & \cdots & 0\\
\vdots & & & & & \vdots\\
0 & 0 & \cdots & 0 & \cdots & 0
\end{pmatrix},
\end{equation}
where the non-zero blocks correspond to the test instance and the
selected representative reference instance.
The matrix $P$ is an orthogonal projection matrix, satisfying
$P^\top=P$ and $P^2=P$.
Accordingly, the standardized test statistic is written as
\begin{equation}
\widetilde T(\bm Y)
=
\frac{T(\bm Y)}{\sigma}
=
\frac{\|P\bm Y\|_2}{\sigma}.
\end{equation}
Under the null hypothesis and in the absence of adaptive selection,
$\widetilde T(\bm Y)$ follows a chi distribution with $d$ degrees of
freedom.
%


\subsection{Naive $p$-value}
\label{sec:naive-p}

We first restate the naive $p$-value introduced in Eq.~\eqref{eq:naive_p} using the vector notation introduced above.
Let $P_{\mathrm{obs}}$ denote the projection matrix determined by the target and representative reference instances selected from the observed data $\bm y$.
A naive analysis treats these selected instances, and hence $P_{\mathrm{obs}}$, as if they had been fixed in advance, independently of the observed data.
Under this hypothetical fixed-selection setting and the null hypothesis,
\begin{align*}
\frac{\|P_{\mathrm{obs}}\bm Y\|_2}{\sigma} \sim \chi_d.
\end{align*}
Accordingly, the naive $p$-value in Eq.~\eqref{eq:naive_p} can be equivalently written as
\begin{align}
p_{\mathrm{naive}} = \mathbb P \left( \chi_d \ge \widetilde T(\bm y) \right),
\end{align}
where $\chi_d$ denotes a chi-distributed random variable with $d$ degrees of freedom.

Importantly, this $\chi_d$ distribution is the null distribution that would be valid if the target and representative reference instances had been specified independently of the observed data.
In the actual procedure, however, both instances are selected adaptively from the data.
Therefore, applying this fixed-selection null distribution without accounting for the selection events generally yields an invalid $p$-value and does not guarantee control of the Type-I error rate at the nominal significance level.


\subsection{Selective $p$-value}

To remove the selection bias, we adopt the framework of
SI~\citep{lee2016exact}.
To describe the data-dependent selection procedures in a unified
manner, we introduce the notion of a \emph{selection map}.
A selection map is a deterministic function that maps the input data
to the outcome of a selection procedure.
For a generic selection map $\mathcal S$,
$\mathcal S(\bm Y)$ denotes the selection outcome obtained from the
random data $\bm Y$, whereas $\mathcal S(\bm y)$ denotes the outcome
observed from the actual data $\bm y$.
The corresponding selection event is
\begin{align*}
\{
\bm Y :
\mathcal S(\bm Y)
=
\mathcal S(\bm y)
\},
\end{align*}
that is, the set of data realizations that produce the same selection
outcome as the observed data.

In our setting, the overall selection procedure consists of
attention-based target selection and adaptive selection of a
representative normal reference.
We denote the corresponding selection maps by
$\mathcal A(\bm Y)$ and $\mathcal M(\bm Y)$, respectively.
The adaptive reference-selection map $\mathcal M$ is defined by first
retrieving the $k$ nearest reference instances and then selecting
their medoid.
The overall selection map is therefore defined as
\begin{equation}
\mathcal E(\bm Y)
=
\left(
\mathcal A(\bm Y),
\mathcal M(\bm Y)
\right).
\label{eq:selection_map}
\end{equation}
SI is performed conditional on the event
\begin{align}
\mathcal E(\bm Y)
=
\mathcal E(\bm y),
\label{eq:selection_event}
\end{align}
which requires the attention-selection outcome and the selected
representative reference to remain identical to those observed from
$\bm y$.
Importantly, the intermediate $k$-nearest-neighbor set is not itself
included in the conditioning event.
It may change as long as the final selected representative reference
remains unchanged.

As is standard in SI, we further condition on a nuisance statistic to
obtain a computationally tractable conditional distribution.
Specifically, we condition on
\begin{equation}
\mathcal Q(\bm Y)
=
\left(
\frac{P\bm Y}{\|P\bm Y\|_2},
\,
(I_{(M_{\mathrm{ref}}+1)d}-P)\bm Y
\right).
\end{equation}
The first component represents the direction of the projected vector
$P\bm Y$, whereas the second component represents the component of
$\bm Y$ orthogonal to the range of $P$.
Its observed value is given by
\begin{equation}
\mathcal Q(\bm y)
=
\left(
\frac{P\bm y}{\|P\bm y\|_2},
\,
(I_{(M_{\mathrm{ref}}+1)d}-P)\bm y
\right).
\end{equation}

Conditioning on
$\mathcal Q(\bm Y)=\mathcal Q(\bm y)$
fixes the direction of the projected component and the orthogonal
component, leaving only the standardized test statistic
$\widetilde T(\bm Y)$ random.
Consequently, the conditional sample space is reduced to a
one-dimensional search path parameterized by
$\widetilde T(\bm Y)$.
The selection event
$\mathcal E(\bm Y)=\mathcal E(\bm y)$
restricts this search path to a feasible region, yielding a truncated
$\chi$ distribution under the null hypothesis.

To characterize the entire truncation region, we employ a parametric
programming approach for selective inference
~\citep{le2021parametric,le2022more}.
Parametric programming systematically explores the one-dimensional
path and identifies all regions that preserve the observed
attention-selection outcome and the selected representative
reference.
Along this path, the $k$-nearest-neighbor set may change.
Regions with a fixed $k$-nearest-neighbor set are used only as local
partitions for characterizing the adaptive reference-selection map,
and regions corresponding to different neighbor sets are combined
whenever they yield the same selected representative reference.
The resulting one-dimensional parameterization and the
characterization of the truncation region are provided in
Appendix~\ref{app:selection_event};
see also
~\citet{taylor2015statistical,lee2016exact,
fithian2014optimal,sugiyama2021more}.

The selective $p$-value is defined as
\begin{equation}
p_{\mathrm{selective}}
=
{\mathbb P}_{{\mathrm H}_0}
\left(
\widetilde T(\bm Y)
\ge
\widetilde T(\bm y)
\;\middle|\;
\mathcal E(\bm Y)=\mathcal E(\bm y),
\,
\mathcal Q(\bm Y)=\mathcal Q(\bm y)
\right).
\end{equation}
Conditioning on the selection event fixes the identity of the
selected representative reference instance.
Because the projection matrix $P$ is uniquely determined by the
selected representative reference instance, $P$ is also fixed under
this conditioning.
Consequently, the truncation region is determined by the selection
event together with the nuisance statistic.
Under the proposed conditioning scheme, the conditional null
distribution of the standardized test statistic can be characterized
analytically.
The following theorem summarizes this result.

\begin{theorem}
Under the null hypothesis, the conditional distribution
\begin{align*}
\widetilde T(\bm Y)
\mid
\mathcal E(\bm Y)=\mathcal E(\bm y),
\,
\mathcal Q(\bm Y)=\mathcal Q(\bm y)
\end{align*}
is a truncated $\chi$ distribution with $d$ degrees of freedom,
where the truncation region is determined by the conditioning events.
\label{theorem:truncated}
\end{theorem}
Therefore, the selective $p$-value can be computed as the upper-tail
probability of the truncated $\chi$ distribution.
The proof of Theorem~\ref{theorem:truncated} is presented in
Appendix~\ref{app:proof_truncated}.

The following theorem establishes the validity of the proposed
selective $p$-value.

\begin{theorem}
For any significance level $\alpha\in(0,1)$,
\begin{align*}
{\mathbb P}_{{\mathrm H}_0}
\left(
p_{\mathrm{selective}}
\le
\alpha
\;\middle|\;
\mathcal E(\bm Y)=\mathcal E(\bm y),
\,
\mathcal Q(\bm Y)=\mathcal Q(\bm y)
\right)
=
\alpha.
\end{align*}
Hence, the proposed selective $p$-value exactly controls the
conditional Type-I error rate under the null hypothesis.
\label{theorem:significance}
\end{theorem}
The proof of Theorem~\ref{theorem:significance} is presented in
Appendix~\ref{app:proof_selectivep}.

\subsection{Selection maps and selection events}
\label{sec:selection_event}

The proposed statistical test is performed after data-dependent
selection of the target instance and its representative normal
reference instance.
Throughout the inference procedure, the trained feature extractor and
attention network are treated as fixed.
We next define the maps involved in these procedures and clarify which
outcomes are included in the conditioning event.

\paragraph{Attention-based target selection.}
The first selection map corresponds to attention-based instance
selection.
For a test bag, the trained ABMIL model computes an attention logit for
each instance.
As described in Section~\ref{sec:mil}, the proposed SI framework
operates directly on these pre-sigmoid attention logits, and only
instances whose attention logits exceed the prespecified threshold
$\tau$ are selected for statistical testing.
For the test instance, this selection map is defined as
\begin{equation}
\mathcal A(\bm Y)
=
\mathbb I
\left\{
f_{\mathrm{att}}
\left(
f_{\mathrm{enc}}
(
\bm X_{\mathrm{test}}
)
\right)
>
\tau
\right\}.
\end{equation}
The threshold $\tau$ is fixed throughout the inference procedure.
In our experiments, $\tau$ is either set to zero or estimated
beforehand from an independent set of negative instances, where the
threshold corresponds to the top $\rho\%$ of the resulting attention
logits.
The independent instances used to determine $\tau$ are disjoint from
both the test and reference instances used for SI.
The observed attention-selection outcome is preserved when
\begin{equation}
\mathcal A(\bm Y)
=
\mathcal A(\bm y).
\end{equation}

\paragraph{$k$-nearest-neighbor retrieval.}
For each selected test instance, the $k$ nearest reference instances
are determined according to the Euclidean distance in the feature
space.
We define the corresponding retrieval map as
\begin{equation}
\mathcal K(\bm Y)
=
\operatorname{Top}_k
\left(
\left\{
\left\|
f_{\mathrm{enc}}
(
\bm X_{\mathrm{test}}
)
-
f_{\mathrm{enc}}
(
\bm X_{\mathrm{ref}}^j
)
\right\|_2^2
\right\}_{j\in[M_{\mathrm{ref}}]}
\right),
\end{equation}
where $\operatorname{Top}_k(\cdot)$ returns the indices corresponding
to the $k$ smallest distances.
The map $\mathcal K(\bm Y)$ is used internally to determine the
representative reference but is not itself included in the
conditioning event.
Accordingly, the $k$-nearest-neighbor set is allowed to change along
the conditional path as long as the final selected representative
reference remains unchanged.

\paragraph{Adaptive reference selection.}
Among the retrieved nearest neighbors, the representative reference
instance is selected as
\begin{equation}
\mathcal M(\bm Y)
=
\arg\min_{j\in\mathcal K(\bm Y)}
\sum_{j'\in\mathcal K(\bm Y)}
\left\|
f_{\mathrm{enc}}
(
\bm X_{\mathrm{ref}}^j
)
-
f_{\mathrm{enc}}
(
\bm X_{\mathrm{ref}}^{j'}
)
\right\|_2^2.
\end{equation}
Thus, $\mathcal M(\bm Y)$ represents the final output of the adaptive
reference-selection procedure, which consists of
$k$-nearest-neighbor retrieval followed by medoid selection.
The observed reference-selection outcome is preserved when
\begin{equation}
\mathcal M(\bm Y)
=
\mathcal M(\bm y),
\end{equation}
which requires that the selected representative reference instance
remain unchanged.

Any ties in $k$-nearest-neighbor retrieval or medoid selection are
resolved using a prespecified deterministic rule.
The selection inequalities in
Appendix~\ref{app:selection_event}
are interpreted consistently with this rule.
A strict inequality is used whenever a tie would favor a competing
index over the corresponding selected index.

The definitions above use distances in the encoded feature space.
The $k$-nearest-neighbor retrieval and medoid selection can also be
performed in the input space by replacing
$f_{\mathrm{enc}}(\bm X)$ with $\bm X$ in the corresponding maps.
The choice of space is fixed before selective inference.
The attention-selection map remains unchanged.

Collectively, the overall selection map is
\begin{equation}
\mathcal E(\bm Y)
=
\left(
\mathcal A(\bm Y),
\mathcal M(\bm Y)
\right).
\end{equation}
Therefore, the observed selection event is
\begin{equation}
\mathcal E(\bm Y)
=
\mathcal E(\bm y).
\end{equation}
This event requires the attention-selection outcome and the selected
representative reference to remain identical to those observed from
$\bm y$.
It does not require
$\mathcal K(\bm Y)=\mathcal K(\bm y)$.
Instead, regions with different $k$-nearest-neighbor sets are retained
whenever they lead to the same observed representative reference.

Combined with the nuisance statistic
$\mathcal Q(\bm Y)=\mathcal Q(\bm y)$,
the conditional sample space reduces to a one-dimensional truncation
region for the standardized test statistic
$\widetilde T(\bm Y)$.
Appendix~\ref{app:selection_event} explicitly characterizes this
region by partitioning the conditional path into local regions with
fixed $k$-nearest-neighbor sets and combining all such regions that
preserve the observed representative reference.
The following section experimentally verifies that the proposed SI
achieves valid Type-I error control while maintaining high statistical
power.

\section{Experiments}
\label{sec:exp}

\subsection{Experimental settings}
\label{sec:exp_setting}

All experiments were conducted at the significance level $\alpha=0.05$.

The proposed method was evaluated in terms of the Type-I error rate and
statistical power.
The Type-I error rate measures whether the hypothesis test correctly
controls the nominal significance level under the null hypothesis.
For the synthetic and MNIST experiments, statistical power measures
the probability of correctly rejecting the null hypothesis under the
alternative hypothesis.
For the real-world WSI experiment, the evaluation metrics are described
in Section~\ref{sec:exp_camelyon}.

We compared the following methods:

\begin{itemize}

    \item \textbf{Proposed}:
    The proposed selective inference method based on parametric
    programming.
    The method conditions on the attention-based target-selection
    outcome and the adaptively selected representative reference,
    and characterizes the entire truncation region consistent with
    these outcomes.
    The intermediate $k$-nearest-neighbor set is not itself included
    in the conditioning event.

    \item \textbf{Ablation 1}:
    An ablation study that ignores the attention-based selection event.
    The selective $p$-value is computed by conditioning only on the
    adaptive reference-selection outcome.
    This ablation evaluates the effect of ignoring the
    data-dependent selection of the target instance.

    \item \textbf{Ablation 2}:
    An ablation study that ignores adaptive reference selection.
    The selective $p$-value is computed by conditioning only on the
    attention-based selection event.
    This ablation evaluates the effect of ignoring the
    data-dependent selection of the representative reference.

    \item \textbf{Over-conditioning (OC)}:
    An ablation study based on over-conditioning.
    Instead of characterizing the entire truncation region using
    parametric programming, this method computes the selective
    $p$-value using only the local truncation region containing the
    observed test statistic.
    Consequently, OC introduces additional conditioning beyond the
    observed attention and representative-reference outcomes and can
    therefore reduce statistical power.

    \item \textbf{Bonferroni}:
    Bonferroni correction for multiple hypothesis testing.
    To account for attention-based selection, we define
    \[
    \lambda_{\mathrm{att}}
    =
    \frac{N_{\mathrm{sample}}}{N_{\mathrm{att}}},
    \]
    where $N_{\mathrm{sample}}$ is the number of sampled instances and
    $N_{\mathrm{att}}$ is the number of high-attention instances selected
    by attention thresholding.
    For each selected target instance, the adaptive reference-selection
    procedure selects one representative reference from
    $M_{\mathrm{ref}}$ reference candidates through
    $k$-nearest-neighbor retrieval followed by medoid selection.
    We therefore use the multiplicity factor
    \[
    \lambda_{\mathrm{att}} M_{\mathrm{ref}}.
    \]
    The Bonferroni-corrected $p$-value is defined as
    \[
    p_{\mathrm{Bonferroni}}
    =
    \min\left\{
    1,\,
    \lambda_{\mathrm{att}}
    M_{\mathrm{ref}}
    p_{\mathrm{naive}}
    \right\}.
    \]
    Here, $\lambda_{\mathrm{att}}$ accounts for attention-based target
    selection, while $M_{\mathrm{ref}}$ accounts for the possible
    choices of the final representative reference.
    We use this Bonferroni-based correction as a comparison baseline.
    For the synthetic and MNIST-based experiments, we evaluate its
    empirical Type-I error rate and statistical power, whereas for
    CAMELYON16 we report rejection proportions as described in
    Section~\ref{sec:exp_camelyon}.
    The correction can be highly conservative when the multiplicity
    factor is large.

    \item \textbf{Naive}:
    A naive hypothesis testing method that ignores all data-dependent
    selection.
    The naive $p$-value is computed directly from the fixed-selection
    null distribution described in Section~\ref{sec:naive-p}.
    We include this method in the synthetic experiments to demonstrate
    the potential failure of statistical inference that ignores the
    adaptive selection procedures.

\end{itemize}

The naive method is included only in the synthetic experiments to
illustrate the effect of ignoring the adaptive selection procedures
and is not used in the MNIST-based or pathological WSI experiments.

\subsection{Synthetic experiments}
\label{sec:exp_synth}

We first evaluated the proposed method on synthetic data, where the ground-truth distribution is known and the Type-I error rate and statistical power can be quantitatively evaluated.

\paragraph{Experimental settings.}
Positive instances were generated from a multivariate Gaussian distribution
$\mathcal N(\bm{1}, \sigma^2 I_d)$,
whereas negative instances were generated from
$\mathcal N(\bm{0}, \sigma^2 I_d)$.
During training, bags consisting of 10 instances were constructed.
A positive bag contained at least one positive instance, while a negative bag consisted entirely of negative instances.

An ABMIL model with a multilayer perceptron (MLP) backbone was trained.
The attention network was trained with a sigmoid activation at its output, followed by the bag-wise softmax normalization used in ABMIL.
During inference, however, we extracted the pre-sigmoid attention logits and regarded an instance as a high-attention instance when its logit was greater than zero.

The input dimension was varied over
$d\in\{16,32,48,64\}$,
and the corresponding feature dimensions were
$d'\in\{4,8,12,16\}$.
The noise variance was varied over
$\sigma^2\in\{0.6,0.8,1.0,1.2,1.4\}$,
and the number of nearest neighbors was
$k\in\{1,3,5,10\}$.
For each hypothesis test, 100 negative reference instances were independently generated using a different random seed.

Since the noise variance was known in the synthetic setting, the true variance used for data generation was employed in the statistical test.
For the Type-I error rate evaluation, negative instances were used as test instances, whereas positive instances were used for the statistical power evaluation.
Because the attention logit of ABMIL depends only on each individual instance, the hypothesis test was performed at the instance level rather than on bags.
For each experiment, hypothesis testing was repeated until 10,000 high-attention instances were collected.

\begin{figure}[tb]
\begin{center}
 \includegraphics[width=0.4\linewidth]{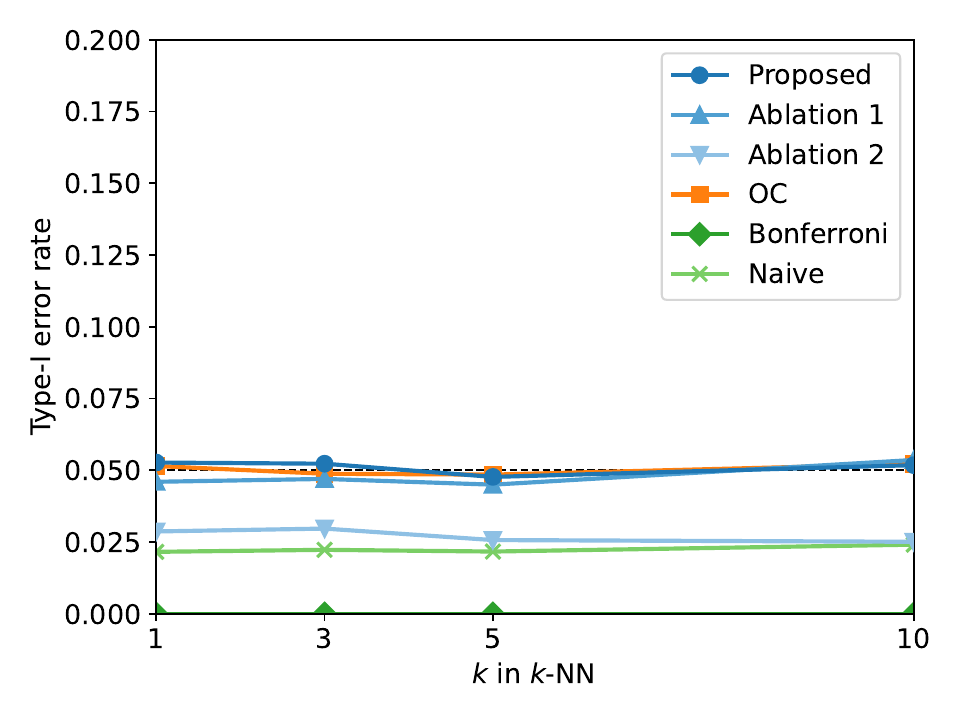} ~~~
 \includegraphics[width=0.4\linewidth]{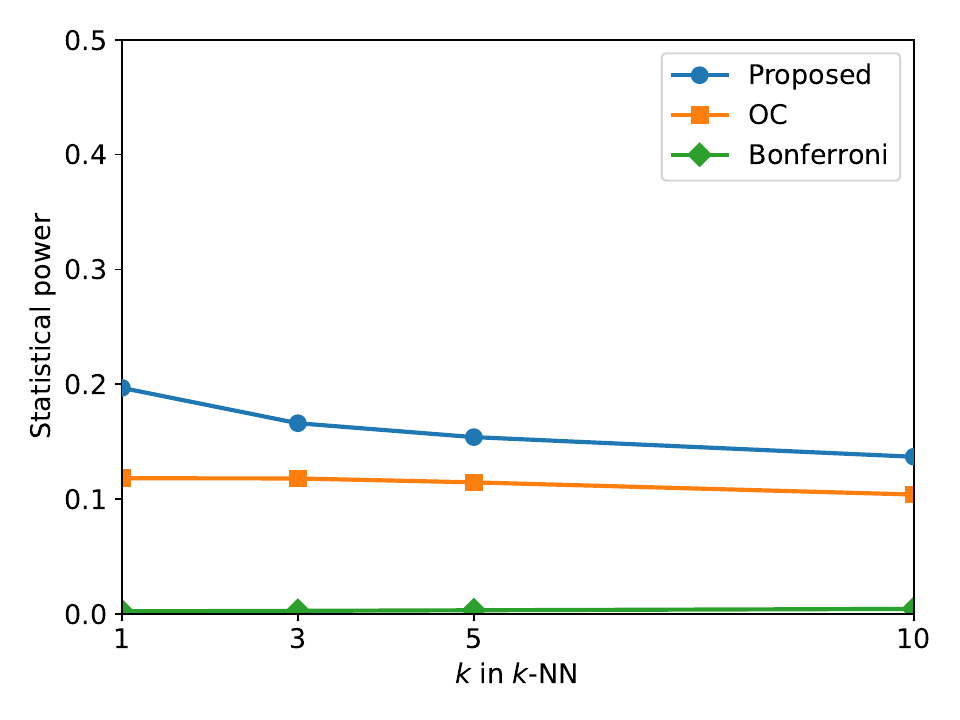} \\
  \includegraphics[width=0.4\linewidth]{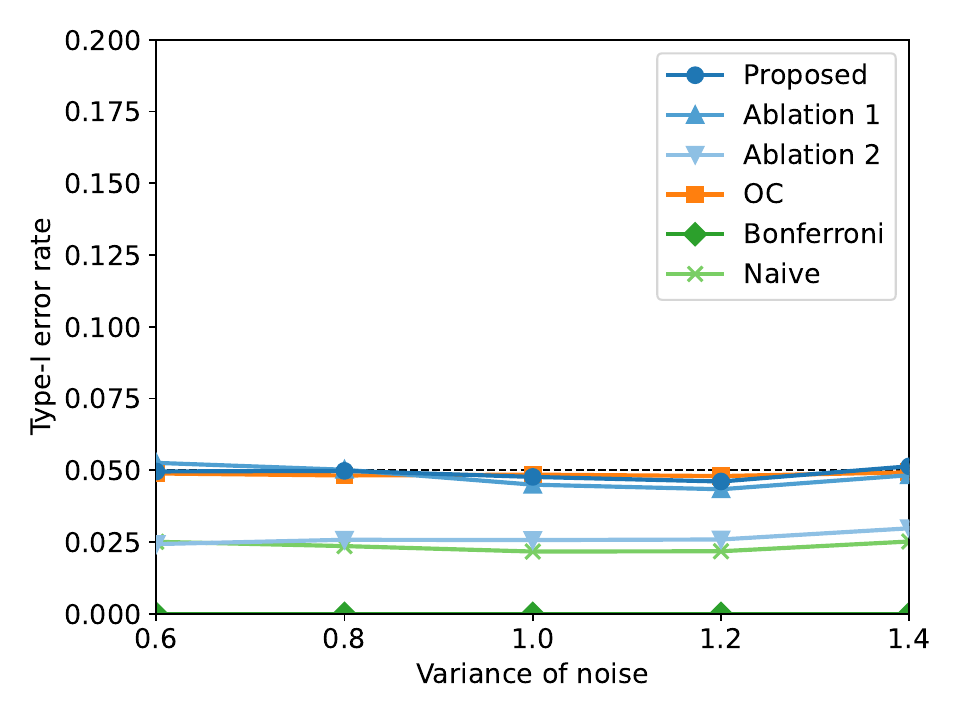} ~~~
  \includegraphics[width=0.4\linewidth]{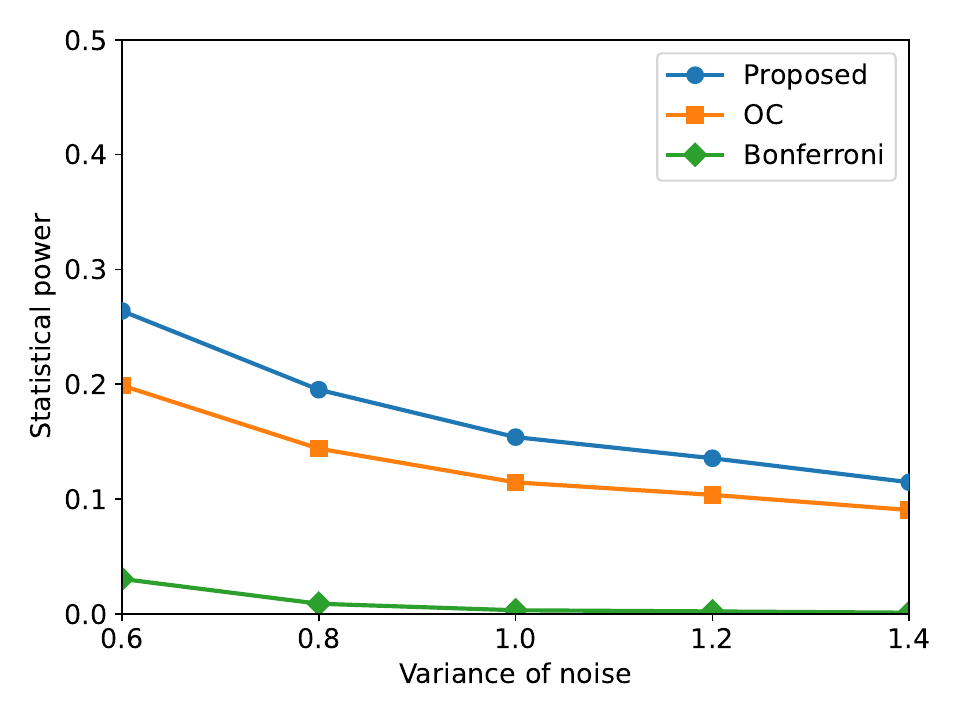}
\end{center}
 \caption{
Type-I error rate (left) and statistical power (right) on the synthetic dataset.
The top row shows the results for varying the number of nearest neighbors $k$, while the bottom row shows the results for varying the Gaussian noise variance.
}
\label{fig:exp_synth}
\end{figure}

\paragraph{Results.}
We evaluated the proposed method and the two ablation methods in terms
of Type-I error control, and compared the statistical power of the
proposed method with those of OC and Bonferroni correction.
Figure~\ref{fig:exp_synth} shows the Type-I error rate and statistical
power obtained by varying the number of nearest neighbors $k$ and the
Gaussian noise variance.
When varying $k$, the input dimension and noise variance were fixed to
$d=32$ and $\sigma^2=1.0$, respectively.
When varying the noise variance, the input dimension and the number of
nearest neighbors were fixed to $d=32$ and $k=5$, respectively.
The proposed method successfully controlled the Type-I error rate at
the nominal significance level under all settings.
Ablation~1 also exhibited Type-I error rates close to the nominal
significance level in the settings considered here, whereas Ablation~2
showed conservative behavior with Type-I error rates substantially
below the nominal level.
The naive method showed a similar conservative tendency to Ablation~2,
suggesting that the bias induced by adaptive reference selection
dominates the relatively weak effect of attention-based selection
under these settings.
OC maintained valid Type-I error control but consistently achieved
lower statistical power than the proposed method, while Bonferroni
correction was highly conservative.
These results demonstrate that the proposed method achieves valid
Type-I error control while providing higher statistical power than OC.
The results for the remaining experimental settings are provided in
Appendix~\ref{app:exp_synth}.

\paragraph{Stress test under strong attention selection.}
We further conducted a stress-test experiment to investigate the effect
of strong attention-based selection on Type-I error control.
Unlike the preceding synthetic experiments, where the attention
threshold was fixed at zero, we used a substantially more stringent
attention threshold.
Specifically, the threshold $\tau$ was set to the 99.99th percentile
of the pre-sigmoid attention logits obtained from an independently
generated set of normal instances.

To isolate the effects of attention-based selection and medoid
selection, we set the number of nearest neighbors equal to the number
of reference instances, i.e., $k=M_{\mathrm{ref}}$.
Under this setting, all reference instances are included in the
neighbor set, so the $k$-nearest-neighbor retrieval becomes trivial,
while medoid selection remains adaptive for $k>1$.
We varied
$M_{\mathrm{ref}}=k\in\{1,3,5,10\}$.
For each setting, negative instances were independently sampled until
1,000 instances satisfying the attention threshold were collected,
and the Type-I error rate was evaluated using these selected instances.

\begin{figure}[tb]
\begin{center}
 \includegraphics[width=0.4\linewidth]
 {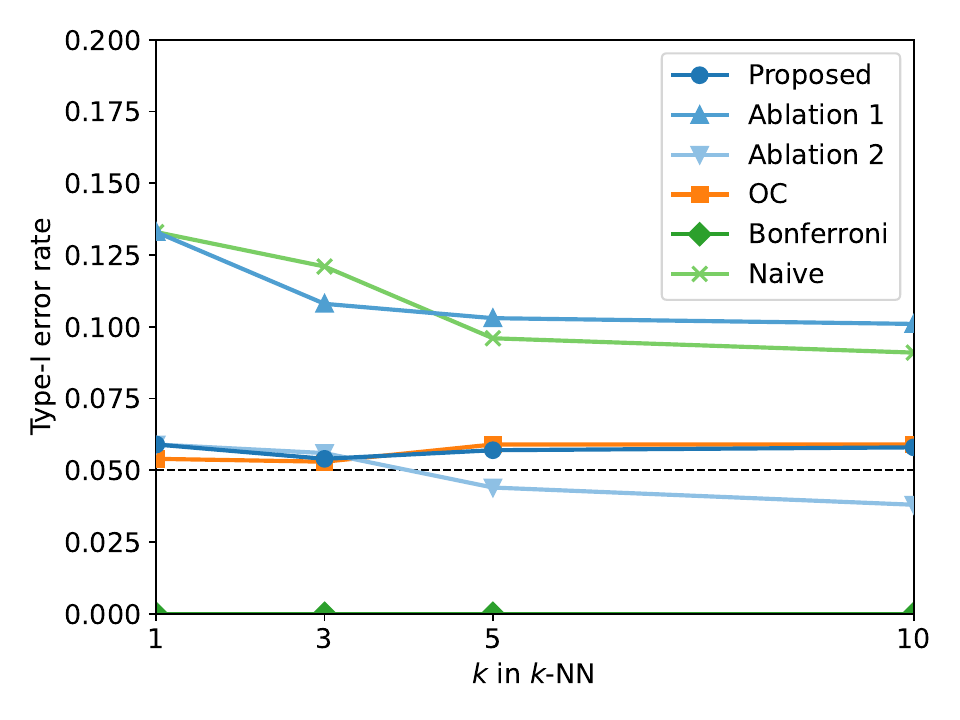}
\end{center}
\caption{
Type-I error rate under strong attention-based selection in the
synthetic stress-test experiment.
The dashed horizontal line indicates the nominal significance level of
0.05.
The number of reference instances and the number of nearest neighbors
were varied jointly as
$M_{\mathrm{ref}}=k\in\{1,3,5,10\}$,
while fixing the input dimension to $d=32$ and the Gaussian noise
variance to $\sigma^2=0.6$.
The attention threshold was set to the 99.99th percentile of the
attention logits obtained from an independent set of normal instances.
}
\label{fig:exp_synth_stress}
\end{figure}

Figure~\ref{fig:exp_synth_stress} shows the Type-I error rates under
strong attention-based selection.
The proposed method maintained the Type-I error rate close to the
nominal significance level across all examined settings.
In contrast, the naive method and Ablation~1, which ignores
attention-based selection, exhibited substantially inflated Type-I
error rates.
Bonferroni correction was highly conservative and yielded almost no
rejections.
These results demonstrate that ignoring attention-based selection can
lead to substantial Type-I error inflation when the selection effect
is strong, whereas the proposed method maintains valid Type-I error
control by explicitly accounting for the adaptive selection process.
Therefore, although the naive method is included in the synthetic
experiments to illustrate the consequence of ignoring the adaptive
selection procedures, we do not consider it a valid baseline in the
subsequent real-data experiments.

\FloatBarrier
\subsection{MNIST-based experiments}
\label{sec:exp_mnist}

To evaluate the proposed method under a more realistic heterogeneous data distribution, we conducted experiments using the MNIST dataset~\citep{deng2012mnist}.
Unlike the synthetic experiments in Section~\ref{sec:exp_synth}, both positive and negative instances were generated from multiple Gaussian clusters whose means were derived from real images.

\paragraph{Experimental settings.}
For the training data, 10 positive and 10 negative images were randomly selected from the training set as cluster centers.
Synthetic feature vectors were then generated by adding Gaussian noise to these images, resulting in 10 Gaussian clusters for each class.
Images of the digit ``0'' were used as the negative class, while each of the digits $\{1,\ldots,9\}$ was evaluated independently as the positive class.

For statistical inference, separate cluster centers were selected from
the MNIST test set.
We considered three- and five-cluster settings.
The same negative cluster centers were used to generate both reference
and negative test instances.
For positive test instances, the same number of positive cluster centers
was independently selected from the corresponding positive digit class.
The selected cluster centers were fixed throughout each experiment,
whereas Gaussian noise was independently regenerated for every
hypothesis test.

Each $28\times28$ image was downsampled to $14\times14$ by average pooling and represented as a 196-dimensional input vector.
This dimensionality reduction was performed because a large input dimension leads to an excessively large degree of freedom for the $\chi$ distribution, resulting in numerical instability during the computation of selective $p$-values.
The feature dimension was fixed to 32, and the noise variance was
fixed to $\sigma^2=0.25$.
The number of reference instances was fixed to 100 regardless of the number of cluster centers, and cluster centers were sampled uniformly at random.

As in the synthetic experiments, the true noise variance was assumed
to be known.
Negative and positive instances were used for the Type-I error rate
and statistical power evaluations, respectively.
For the Type-I error rate evaluation, only hypothesis tests in which
the input instance and the selected representative reference instance
originated from the same Gaussian cluster were counted, ensuring that
both samples were generated from the same underlying distribution
under the null hypothesis.
To introduce a nontrivial attention-based selection effect, the
attention threshold was determined using an independent set of
negative instances.
Specifically, $\tau$ was set to the empirical quantile corresponding
to the top $5\%$ of their pre-sigmoid attention logits.
These negative instances were independent of the test and reference
instances used for selective inference, and the resulting threshold
was treated as fixed throughout the inference procedure.
For Type-I error evaluation, sampling continued until 10,000
high-attention test instances whose selected references originated
from the same negative cluster were obtained.

\begin{figure}[tb]
\begin{center}
 \includegraphics[width=0.95\linewidth]{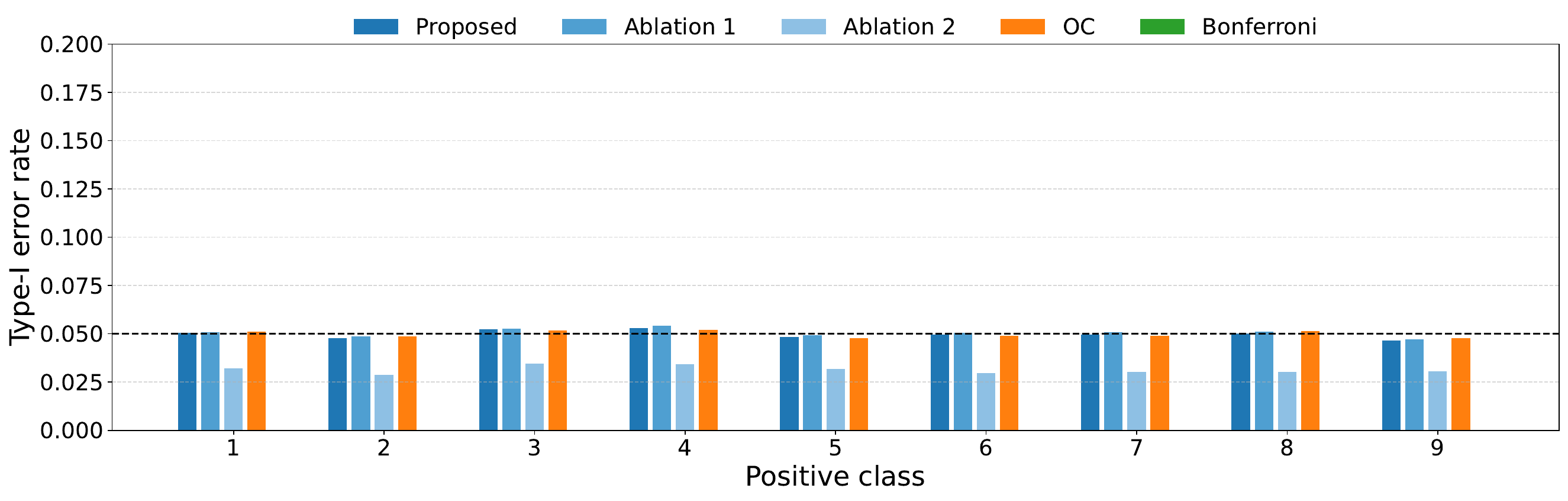}\\
 \includegraphics[width=0.95\linewidth]{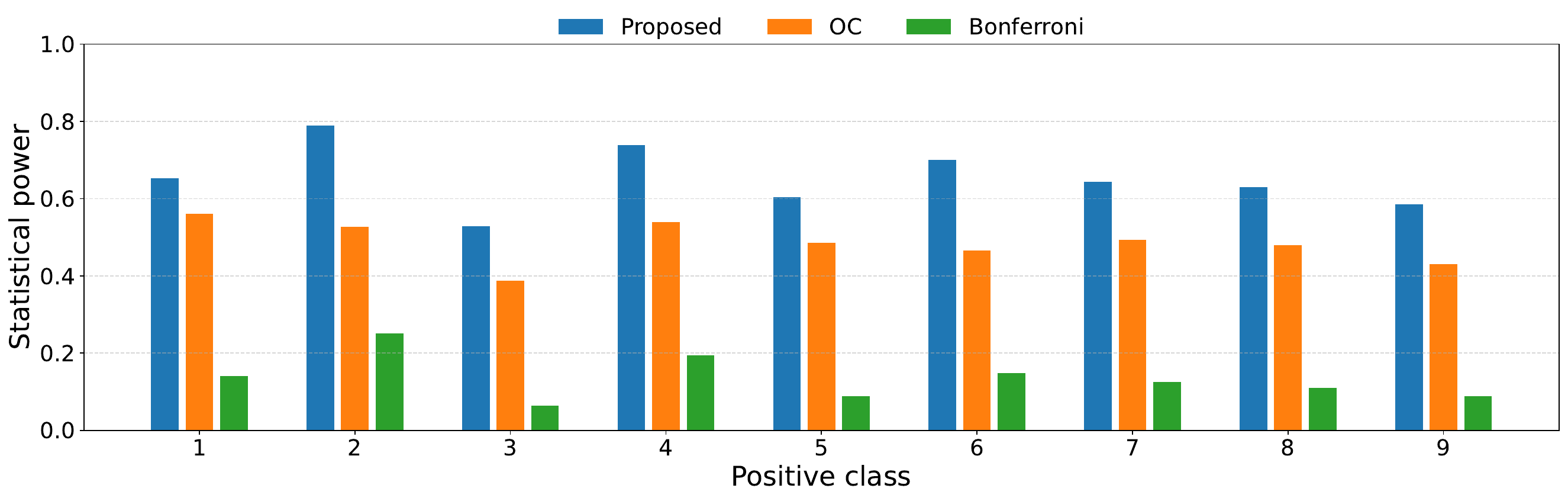}\\
 \includegraphics[width=0.95\linewidth]{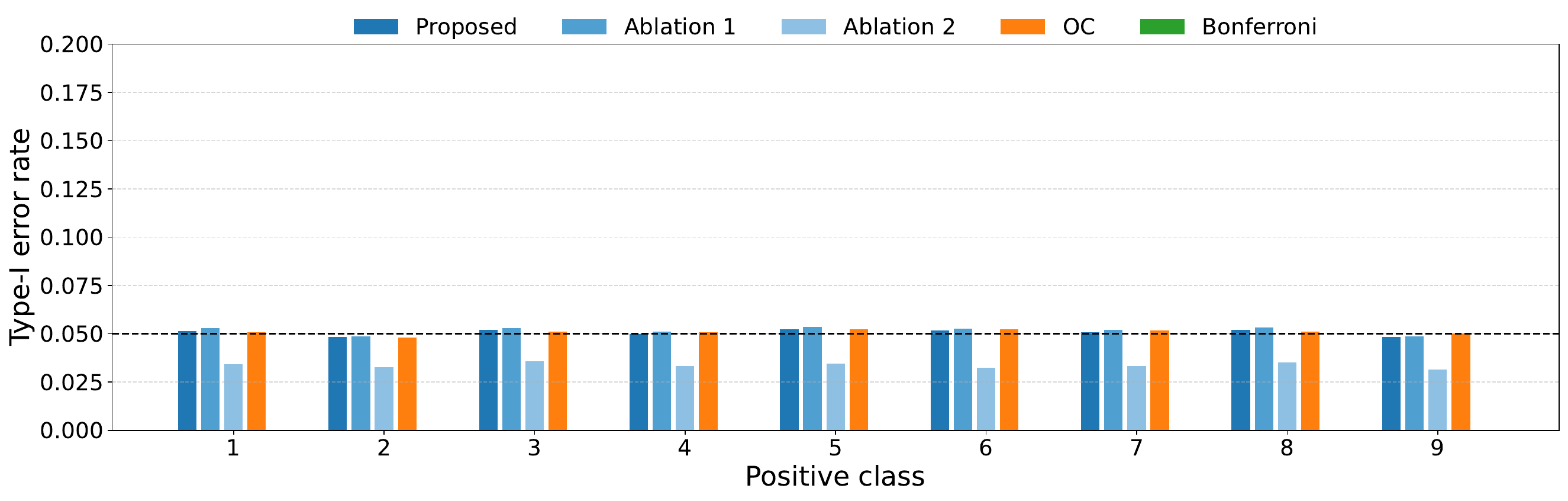}\\
 \includegraphics[width=0.95\linewidth]{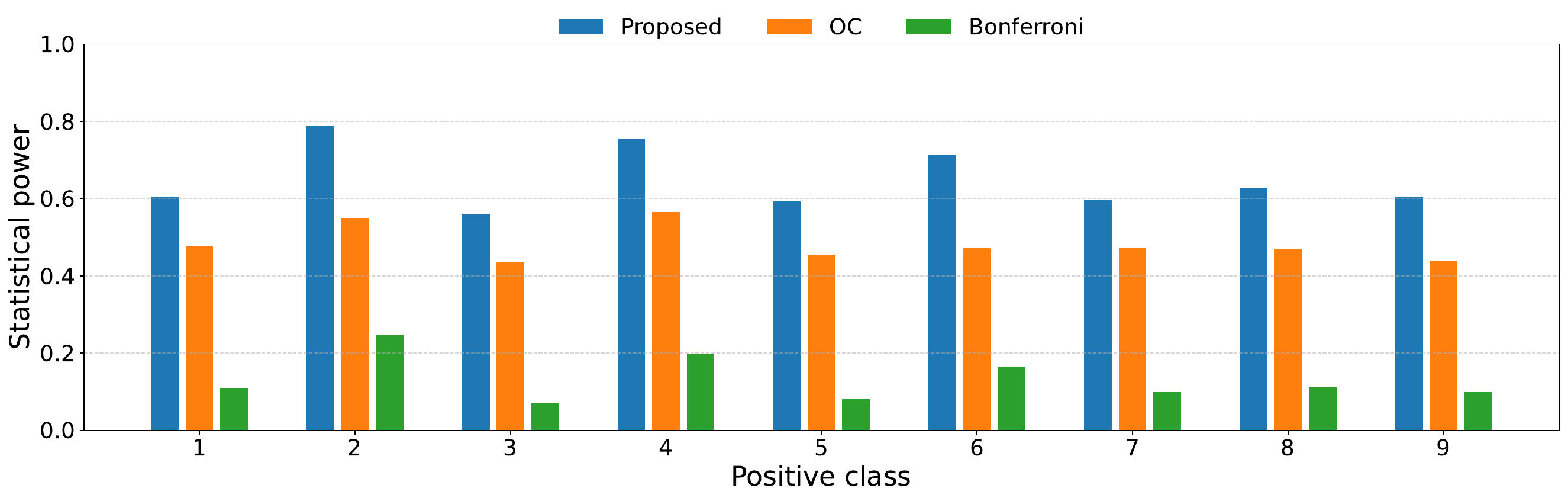}
\end{center}
 \caption{
Type-I error rate and statistical power in the MNIST-based experiments.
From top to bottom, the plots show the Type-I error rate and statistical power using three Gaussian clusters, followed by the Type-I error rate and statistical power using five Gaussian clusters.
The horizontal axis represents the positive class.
}
\label{fig:exp_mnist}
\end{figure}

\paragraph{Results.}
Figure~\ref{fig:exp_mnist} shows the Type-I error rate and statistical power for $k=5$ and $\sigma^2=0.25$, where each digit from 1 to 9 is treated as the positive class.
The top and bottom panels correspond to the settings with three and five Gaussian clusters, respectively.
The proposed method successfully controlled the Type-I error rate at
the nominal significance level across all classes in both settings.
Ablation~1 achieved a Type-I error rate close to the nominal level,
suggesting that the attention-based selection event has a relatively
small impact in this setting.
In contrast, Ablation~2 exhibited conservative behavior in the
examined settings, illustrating the effect of ignoring adaptive
reference selection.
Compared with OC, the proposed method consistently achieved higher
statistical power while maintaining valid Type-I error control,
whereas Bonferroni correction was overly conservative.
Additional results for other values of $k$ are provided in
Appendix~\ref{app:exp_mnist}.

\subsection{Real-world experiments on CAMELYON16}
\label{sec:exp_camelyon}

Finally, we evaluated the proposed method on the CAMELYON16 dataset~\citep{bejnordi2017diagnostic} to demonstrate its applicability to a realistic digital pathology setting.
Unlike the previous experiments, this experiment was conducted on WSIs using an ABMIL framework.

\paragraph{Experimental settings.}
The CAMELYON16 dataset contains WSIs collected at two medical centers and includes both micro- and macro-metastatic tumor cases.
In this experiment, we used only the WSIs collected at one of the two centers.
This restriction was introduced to reduce center-specific variations arising from differences in staining and slide preparation.
Although this experiment considers heterogeneous reference distributions, our primary interest is the heterogeneity among normal tissue samples.
We therefore restricted the analysis to a single center to avoid introducing additional sources of variation and to keep the experimental setting as simple as possible.
We also included only macro-metastatic tumor cases because their relatively large tumor regions facilitate the interpretation of the visualization results.

The training split from the selected center contained 59 normal WSIs
and 26 macro-metastatic tumor WSIs.
Among the 59 normal WSIs, 37 were used for model training, 11 were
reserved for reference construction, and the remaining 11 were allocated exclusively for noise-variance estimation and attention-threshold
determination.
All 26 macro-metastatic tumor WSIs were used for model training.
These three subsets of normal WSIs were mutually disjoint.
The test set consisted of 30 normal WSIs and 9 macro-metastatic tumor
WSIs.

Each WSI was divided into non-overlapping $2048\times2048$-pixel patches over its tissue regions.
Each patch was resized to the input resolution required by the UNI2 foundation model~\citep{chen2024towards}, from which a 1,536-dimensional feature vector was extracted.
To reduce the degrees of freedom of the $\chi$ distribution and avoid numerical instability in the computation of selective $p$-values, average pooling was applied to reduce the UNI2 feature dimension from 1,536 to 192.
The resulting 192-dimensional vectors were regarded as input samples and were used for both hypothesis testing and reference selection.
Unlike the synthetic and MNIST experiments, the pathology classification task is substantially more complex, and the neighborhood relationships in the input space may not be well preserved in the intermediate feature space learned by the neural network.
To ensure that the selected medoid remains a representative sample of
the original input instances, the $k$-nearest-neighbor search was
therefore performed in the 192-dimensional input space rather than in
the learned feature space used in
Section~\ref{sec:selection_event}.
Although the attention-based MIL model contained a 32-dimensional intermediate feature representation, this representation was not used for reference selection in this experiment.

The reference instances were constructed from the 11 normal WSIs
reserved for reference construction.
From each reference WSI, up to 100 input samples were randomly selected.
Since several WSIs contained fewer than 100 eligible patches, the
reference set comprised 1,009 samples in total.
Each reference WSI was regarded as a distinct population with a
slide-specific mean vector, thereby accounting for heterogeneity among
normal tissue samples.
The $k$-nearest-neighbor search was performed in the
192-dimensional input space with $k=3$.

The remaining 11 normal WSIs reserved for noise-variance estimation
were also used to determine the attention thresholds.
From each WSI, up to 100 input samples were randomly selected.
Each WSI was modeled as an isotropic Gaussian distribution with its own
mean vector.
The isotropic variance was estimated independently for each WSI, and
the final noise variance was obtained by averaging the estimated
variances over the 11 WSIs.
The empirical distribution of the pre-sigmoid attention logits from
these independently sampled patches was used to determine the threshold
corresponding to the top $5\%$ of attention logits.
Both the estimated noise variance and the attention thresholds were
treated as fixed throughout selective inference.

Because the number of instances varies across WSIs, the multiplicity
factor for attention-based selection was computed separately for each
WSI.
Specifically, for the $n$-th WSI, we set
\[
\lambda_{\mathrm{att}}^{n}
=
\frac{M_{\mathrm{test}}^{n}}
{|\mathcal C(X_{\mathrm{test}}^{n})|},
\]
where $M_{\mathrm{test}}^{n}$ and
$|\mathcal C(X_{\mathrm{test}}^{n})|$ denote the total number of
instances and the number of high-attention instances in the WSI,
respectively.
Accordingly, the WSI-specific Bonferroni multiplicity factor was
\[
\lambda_{\mathrm{att}}^{n} M_{\mathrm{ref}}.
\]

\paragraph{Results.}

Table~\ref{tab:exp_cam_eval} summarizes the results obtained with the
attention threshold fixed to the top $5\%$ of the independent
threshold-estimation instances and $k=3$.

For normal WSIs, we report the rejection proportion among
attention-selected patches as an empirical proxy for the
Type-I error rate.
For tumor WSIs, we report the corresponding rejection proportion
as an empirical proxy for statistical power.
Because the equality or inequality of the underlying signal vectors
is not directly observed for individual patch pairs, these quantities
are operational evaluation metrics.

The proposed method yielded a rejection proportion close to the
nominal significance level on normal WSIs while achieving a
substantially higher rejection proportion than OC on tumor WSIs.
Bonferroni yielded low rejection proportions on both normal and
tumor WSIs.

These quantitative results demonstrate that the proposed selective
inference framework remains applicable in a realistic digital
pathology setting despite the heterogeneity of normal tissue
distributions.

\begin{figure*}[tb]
\begin{center}
\includegraphics[width=0.97\linewidth]{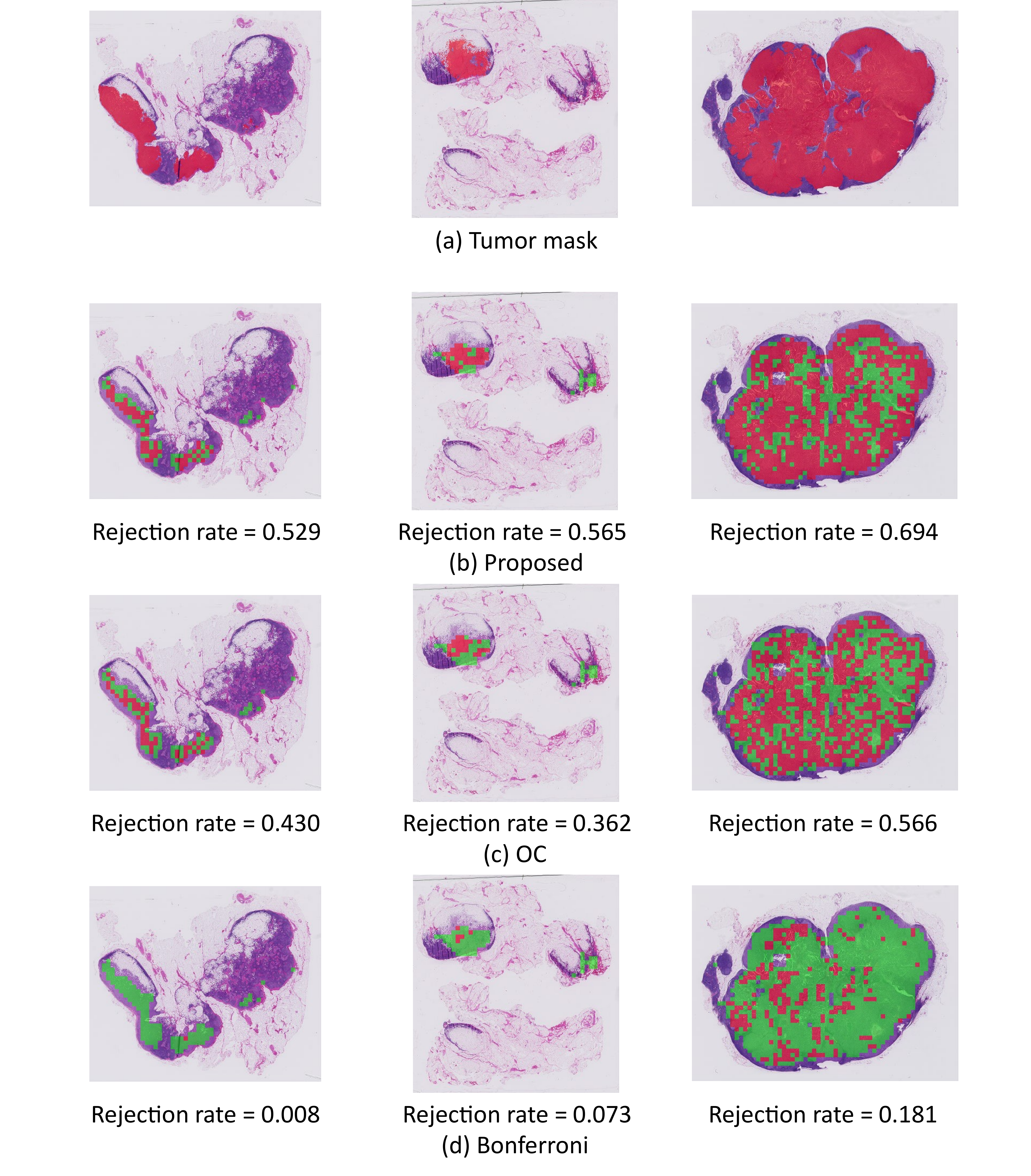}
\end{center}
 \caption{
Visualization of $p$-value maps for three representative
CAMELYON16 tumor WSIs.
(a) shows the ground-truth tumor masks, while (b), (c), and (d) show the results of Proposed, OC, and Bonferroni, respectively.
Green and red rectangles indicate selected high-attention instances with $p \geq 0.05$ and $p < 0.05$, respectively.
The remaining tumor cases are provided in Appendix~\ref{app:exp_camelyon}.
}
\label{fig:exp_cam_vis}
\end{figure*}

Figure~\ref{fig:exp_cam_vis} presents representative visualization results on three tumor WSIs.
The colored squares indicate patches selected by the attention mechanism and subjected to hypothesis testing.
Red squares denote rejected hypotheses, whereas green squares denote non-rejected hypotheses.
The tumor masks are provided only for visualization and were not used during inference.
Additional qualitative results are
provided in Appendix~\ref{app:exp_camelyon}.

Although both methods performed hypothesis testing on the same
attention-selected patches, the proposed method rejected
substantially more hypotheses than OC.
Consequently, more attention-selected patches were identified as statistically significant.
In contrast, Bonferroni correction rejected substantially fewer
patches than the proposed method.
As shown in the second column of Fig.~\ref{fig:exp_cam_vis}, the tissue region on the right
side of the slide receives a high attention logit and is therefore
selected for statistical testing.
However, the proposed test does not detect a significant difference
between this region and the selected reference instance.
This example illustrates that a high attention logit alone does not
necessarily imply a statistically significant deviation from the
selected normal reference.
These qualitative observations agree with the quantitative results
in Table~\ref{tab:exp_cam_eval}, demonstrating that the proposed
method improves the identification of statistically significant
regions.

\begin{table}[tb]
\caption{
Quantitative evaluation on the CAMELYON16 dataset for $k=3$.
The numbers in parentheses indicate the numbers of tested instances
used to compute the corresponding metrics.
}
    \label{tab:exp_cam_eval}
  \centering
    \begin{tabular}{crrr}
  \hline
  & \multicolumn{1}{c}{Proposed} & \multicolumn{1}{c}{OC} & \multicolumn{1}{c}{Bonferroni} \\ \hline
Normal rejection rate (297) & 0.037 & 0.037 & 0.000  \\ 
Tumor rejection rate (1962) & \underline{0.570} & 0.468 & 0.117 \\ \hline
  \end{tabular}
\end{table}

\FloatBarrier
\section{Conclusion}

In this paper, we proposed a selective inference framework for attention-based multiple instance learning with adaptive reference selection.
The proposed method accounts for attention-based target selection and
adaptive reference selection, in which the representative reference is
selected through $k$-nearest-neighbor retrieval followed by medoid
selection.
Importantly, the intermediate $k$-nearest-neighbor set is not itself
conditioned upon; instead, the proposed framework conditions on the
final selected representative reference.
We further developed a parametric-programming-based procedure that
computes valid selective $p$-values by characterizing the entire
truncation region consistent with the observed selection outcome.

Experiments on synthetic and MNIST-based data demonstrated Type-I error
control and higher statistical power than the conventional
over-conditioning approach.
The CAMELYON16 experiment further demonstrated the practical
applicability of the proposed framework, with a higher rejection
proportion than over-conditioning on tumor WSIs.
These results indicate that the proposed framework provides a
statistically principled approach for evaluating attention-selected
instances through comparison with adaptively selected reference
instances.

One limitation of the present work is that we demonstrated the proposed
framework using ABMIL as a proof-of-concept architecture.
ABMIL provides a relatively simple attention mechanism that allows the
data-dependent selection procedure to be explicitly characterized.
More generally, the proposed framework is applicable when the
data-dependent selection procedure can be characterized along the
one-dimensional conditional path.
In particular, neural networks composed of affine operations and
piecewise-linear activations such as ReLU are piecewise affine with
respect to the input, facilitating characterization of the
corresponding selection constraints by parametric programming.

Future work includes extending the proposed framework to modern MIL
architectures and foundation-model-based pathology
models~\citep{chen2024towards,vorontsov2024foundation,ding2025multimodal}
while preserving statistical validity.
We also plan to investigate selective inference for a broader range of
retrieval and explanation mechanisms used in digital pathology.


\subsubsection*{Acknowledgments}
This work was partially supported by MEXT KAKENHI (24K15080, 26H02528, 26K15711), JST CREST (JPMJCR21D3, JPMJCR22N2), and RIKEN Center for Advanced Intelligence Project.

\clearpage

\bibliographystyle{unsrtnat}
\bibliography{main}

@inproceedings{jain2019attention,
  title={Attention is not explanation},
  author={Jain, Sarthak and Wallace, Byron C},
  booktitle={Proceedings of the 2019 Conference of the North American Chapter of the Association for Computational Linguistics: Human Language Technologies, Volume 1 (Long and Short Papers)},
  pages={3543--3556},
  year={2019}
}

@article{kriegeskorte2009circular,
  title={Circular analysis in systems neuroscience: the dangers of double dipping},
  author={Kriegeskorte, Nikolaus and Simmons, W Kyle and Bellgowan, Patrick SF and Baker, Chris I},
  journal={Nature Neuroscience},
  volume={12},
  number={5},
  pages={535--540},
  year={2009},
  doi={10.1038/nn.2303}
}

@article{lu2021data,
  title={Data-efficient and weakly supervised computational pathology on whole-slide images},
  author={Lu, Ming Y and Williamson, Drew FK and Chen, Tiffany Y and Chen, Richard J and Barbieri, Matteo and Mahmood, Faisal},
  journal={Nature biomedical engineering},
  volume={5},
  number={6},
  pages={555--570},
  year={2021},
  publisher={Nature Publishing Group UK London}
}

@inproceedings{serrano2019attention,
  title={Is attention interpretable?},
  author={Serrano, Sofia and Smith, Noah A},
  booktitle={Proceedings of the 57th annual meeting of the association for computational linguistics},
  pages={2931--2951},
  year={2019}
}

@inproceedings{wiegreffe2019attention,
  title={Attention is not not explanation},
  author={Wiegreffe, Sarah and Pinter, Yuval},
  booktitle={Proceedings of the 2019 conference on empirical methods in natural language processing and the 9th international joint conference on natural language processing (EMNLP-IJCNLP)},
  pages={11--20},
  year={2019}
}

@article{wang2022label,
  title={Label cleaning multiple instance learning: Refining coarse annotations on single whole-slide images},
  author={Wang, Zhenzhen and Saoud, Carla and Wangsiricharoen, Sintawat and James, Aaron W and Popel, Aleksander S and Sulam, Jeremias},
  journal={IEEE transactions on medical imaging},
  volume={41},
  number={12},
  pages={3952--3968},
  year={2022},
  publisher={IEEE}
}

@article{sun2025label,
  title={Label-free Concept Based Multiple Instance Learning for Gigapixel Histopathology},
  author={Sun, Susu and Tessier, Leslie and Meeuwsen, Fr{\'e}d{\'e}rique and Grisi, Cl{\'e}ment and van Midden, Dominique and Litjens, Geert and Baumgartner, Christian F},
  journal={arXiv preprint arXiv:2501.02922},
  year={2025}
}

@article{cai2024rethinking,
  title={Rethinking attention-based multiple instance learning for whole-slide pathological image classification: An instance attribute viewpoint},
  author={Cai, Linghan and Huang, Shenjin and Zhang, Ye and Lu, Jinpeng and Zhang, Yongbing},
  journal={arXiv preprint arXiv:2404.00351},
  year={2024}
}

@article{zehnder2025diagnostic,
  title={Diagnostic classification in toxicologic pathology using attention-guided weak supervision and whole slide image features: a pilot study in rat livers},
  author={Zehnder, Philip and Feng, Jeffrey and Nguyen, Trung and Shen, Philip and Sullivan, Ruth and Fuji, Reina N and Hu, Fangyao},
  journal={Scientific Reports},
  volume={15},
  number={1},
  pages={4202},
  year={2025},
  publisher={Nature Publishing Group UK London}
}

@article{dietterich1997solving,
  title={Solving the multiple instance problem with axis-parallel rectangles},
  author={Dietterich, Thomas G and Lathrop, Richard H and Lozano-P{\'e}rez, Tom{\'a}s},
  journal={Artificial intelligence},
  volume={89},
  number={1-2},
  pages={31--71},
  year={1997},
  publisher={Elsevier}
}

@article{maron1997framework,
  title={A framework for multiple-instance learning},
  author={Maron, Oded and Lozano-P{\'e}rez, Tom{\'a}s},
  journal={Advances in neural information processing systems},
  volume={10},
  year={1997}
}

@inproceedings{zhou2002neural,
  title={Neural networks for multi-instance learning},
  author={Zhou, Zhi-Hua and Zhang, Min-Ling},
  booktitle={Proceedings of the International Conference on Intelligent Information Technology, Beijing, China},
  pages={455--459},
  year={2002},
  organization={Citeseer}
}

@article{andrews2002support,
  title={Support vector machines for multiple-instance learning},
  author={Andrews, Stuart and Tsochantaridis, Ioannis and Hofmann, Thomas},
  journal={Advances in neural information processing systems},
  volume={15},
  year={2002}
}

@article{deng2012mnist,
  title={The mnist database of handwritten digit images for machine learning research [best of the web]},
  author={Deng, Li},
  journal={IEEE signal processing magazine},
  volume={29},
  number={6},
  pages={141--142},
  year={2012},
  publisher={IEEE}
}

@inproceedings{cruz2014automatic,
  title={Automatic detection of invasive ductal carcinoma in whole slide images with convolutional neural networks},
  author={Cruz-Roa, Angel and Basavanhally, Ajay and Gonz{\'a}lez, Fabio and Gilmore, Hannah and Feldman, Michael and Ganesan, Shridar and Shih, Natalie and Tomaszewski, John and Madabhushi, Anant},
  booktitle={Medical Imaging 2014: Digital Pathology},
  volume={9041},
  pages={904103},
  year={2014},
  organization={SPIE}
}

@inproceedings{wu2015deep,
  title={Deep multiple instance learning for image classification and auto-annotation},
  author={Wu, Jiajun and Yu, Yinan and Huang, Chang and Yu, Kai},
  booktitle={Proceedings of the IEEE conference on computer vision and pattern recognition},
  pages={3460--3469},
  year={2015}
}

@article{shrivastava2015generalized,
  title={Generalized dictionaries for multiple instance learning},
  author={Shrivastava, Ashish and Patel, Vishal M and Pillai, Jaishanker K and Chellappa, Rama},
  journal={International Journal of Computer Vision},
  volume={114},
  pages={288--305},
  year={2015},
  publisher={Springer}
}

@article{kraus2016classifying,
  title={Classifying and segmenting microscopy images with deep multiple instance learning},
  author={Kraus, Oren Z and Ba, Jimmy Lei and Frey, Brendan J},
  journal={Bioinformatics},
  volume={32},
  number={12},
  pages={i52--i59},
  year={2016},
  publisher={Oxford University Press}
}

@inproceedings{feng2017deep,
  title={Deep MIML network},
  author={Feng, Ji and Zhou, Zhi-Hua},
  booktitle={Proceedings of the AAAI conference on artificial intelligence},
  year={2017}
}

@article{vaswani2017attention,
  title={Attention is all you need},
  author={Vaswani, Ashish and Shazeer, Noam and Parmar, Niki and Uszkoreit, Jakob and Jones, Llion and Gomez, Aidan N and Kaiser, {\L}ukasz and Polosukhin, Illia},
  journal={Advances in neural information processing systems},
  volume={30},
  year={2017}
}

@article{bejnordi2017diagnostic,
  title={Diagnostic assessment of deep learning algorithms for detection of lymph node metastases in women with breast cancer},
  author={Bejnordi, Babak Ehteshami and Veta, Mitko and Van Diest, Paul Johannes and Van Ginneken, Bram and Karssemeijer, Nico and Litjens, Geert and Van Der Laak, Jeroen AWM and Hermsen, Meyke and Manson, Quirine F and Balkenhol, Maschenka and others},
  journal={Jama},
  volume={318},
  number={22},
  pages={2199--2210},
  year={2017},
  publisher={American Medical Association}
}

@inproceedings{ilse2018attention,
  title={Attention-based deep multiple instance learning},
  author={Ilse, Maximilian and Tomczak, Jakub and Welling, Max},
  booktitle={International conference on machine learning},
  pages={2127--2136},
  year={2018},
  organization={PMLR}
}

@article{wang2018revisiting,
  title={Revisiting multiple instance neural networks},
  author={Wang, Xinggang and Yan, Yongluan and Tang, Peng and Bai, Xiang and Liu, Wenyu},
  journal={Pattern Recognition},
  volume={74},
  pages={15--24},
  year={2018},
  publisher={Elsevier}
}

@inproceedings{li2018thoracic,
  title={Thoracic disease identification and localization with limited supervision},
  author={Li, Zhe and Wang, Chong and Han, Mei and Xue, Yuan and Wei, Wei and Li, Li-Jia and Fei-Fei, Li},
  booktitle={Proceedings of the IEEE conference on computer vision and pattern recognition},
  pages={8290--8299},
  year={2018}
}

@inproceedings{das2018multiple,
  title={Multiple instance learning of deep convolutional neural networks for breast histopathology whole slide classification},
  author={Das, Kausik and Conjeti, Sailesh and Roy, Abhijit Guha and Chatterjee, Jyotirmoy and Sheet, Debdoot},
  booktitle={2018 IEEE 15th International Symposium on Biomedical Imaging (ISBI 2018)},
  pages={578--581},
  year={2018},
  organization={IEEE}
}

@inproceedings{lee2019set,
  title={Set transformer: A framework for attention-based permutation-invariant neural networks},
  author={Lee, Juho and Lee, Yoonho and Kim, Jungtaek and Kosiorek, Adam and Choi, Seungjin and Teh, Yee Whye},
  booktitle={International conference on machine learning},
  pages={3744--3753},
  year={2019},
  organization={PMLR}
}

@article{sudharshan2019multiple,
  title={Multiple instance learning for histopathological breast cancer image classification},
  author={Sudharshan, PJ and Petitjean, Caroline and Spanhol, Fabio and Oliveira, Luiz Eduardo and Heutte, Laurent and Honeine, Paul},
  journal={Expert Systems with Applications},
  volume={117},
  pages={103--111},
  year={2019},
  publisher={Elsevier}
}

@article{campanella2019clinical,
  title={Clinical-grade computational pathology using weakly supervised deep learning on whole slide images},
  author={Campanella, Gabriele and Hanna, Matthew G and Geneslaw, Luke and Miraflor, Allen and Werneck Krauss Silva, Vitor and Busam, Klaus J and Brogi, Edi and Reuter, Victor E and Klimstra, David S and Fuchs, Thomas J},
  journal={Nature medicine},
  volume={25},
  number={8},
  pages={1301--1309},
  year={2019},
  publisher={Nature Publishing Group US New York}
}

@article{yamamoto2019automated,
  title={Automated acquisition of explainable knowledge from unannotated histopathology images},
  author={Yamamoto, Yoichiro and Tsuzuki, Toyonori and Akatsuka, Jun and Ueki, Masao and Morikawa, Hiromu and Numata, Yasushi and Takahara, Taishi and Tsuyuki, Takuji and Tsutsumi, Kotaro and Nakazawa, Ryuto and others},
  journal={Nature communications},
  volume={10},
  number={1},
  pages={5642},
  year={2019},
  publisher={Nature Publishing Group UK London}
}

@article{chen2024towards,
  title={Towards a general-purpose foundation model for computational pathology},
  author={Chen, Richard J and Ding, Tong and Lu, Ming Y and Williamson, Drew FK and Jaume, Guillaume and Song, Andrew H and Chen, Bowen and Zhang, Andrew and Shao, Daniel and Shaban, Muhammad and others},
  journal={Nature medicine},
  volume={30},
  number={3},
  pages={850--862},
  year={2024},
  publisher={Nature Publishing Group US New York}
}

@inproceedings{hashimoto2020multi,
  title={Multi-scale domain-adversarial multiple-instance CNN for cancer subtype classification with unannotated histopathological images},
  author={Hashimoto, Noriaki and Fukushima, Daisuke and Koga, Ryoichi and Takagi, Yusuke and Ko, Kaho and Kohno, Kei and Nakaguro, Masato and Nakamura, Shigeo and Hontani, Hidekata and Takeuchi, Ichiro},
  booktitle={Proceedings of the IEEE/CVF conference on computer vision and pattern recognition},
  pages={3852--3861},
  year={2020}
}

@inproceedings{chen2021multimodal,
  title={Multimodal co-attention transformer for survival prediction in gigapixel whole slide images},
  author={Chen, Richard J and Lu, Ming Y and Weng, Wei-Hung and Chen, Tiffany Y and Williamson, Drew FK and Manz, Trevor and Shady, Maha and Mahmood, Faisal},
  booktitle={Proceedings of the IEEE/CVF International Conference on Computer Vision},
  pages={4015--4025},
  year={2021}
}

@article{shao2021transmil,
  title={Transmil: Transformer based correlated multiple instance learning for whole slide image classification},
  author={Shao, Zhuchen and Bian, Hao and Chen, Yang and Wang, Yifeng and Zhang, Jian and Ji, Xiangyang and others},
  journal={Advances in neural information processing systems},
  volume={34},
  pages={2136--2147},
  year={2021}
}

@inproceedings{li2021dual,
  title={Dual-stream multiple instance learning network for whole slide image classification with self-supervised contrastive learning},
  author={Li, Bin and Li, Yin and Eliceiri, Kevin W},
  booktitle={Proceedings of the IEEE/CVF conference on computer vision and pattern recognition},
  pages={14318--14328},
  year={2021}
}

@inproceedings{chen2022scaling,
  title={Scaling vision transformers to gigapixel images via hierarchical self-supervised learning},
  author={Chen, Richard J and Chen, Chengkuan and Li, Yicong and Chen, Tiffany Y and Trister, Andrew D and Krishnan, Rahul G and Mahmood, Faisal},
  booktitle={Proceedings of the IEEE/CVF Conference on Computer Vision and Pattern Recognition},
  pages={16144--16155},
  year={2022}
}

@inproceedings{zhang2022dtfd,
  title={DTFD-MIL: Double-tier feature distillation multiple instance learning for histopathology whole slide image classification},
  author={Zhang, Hongrun and Meng, Yanda and Zhao, Yitian and Qiao, Yihong and Yang, Xiaoyun and Coupland, Sarah E and Zheng, Yalin},
  booktitle={Proceedings of the IEEE/CVF Conference on Computer Vision and Pattern Recognition},
  pages={18802--18812},
  year={2022}
}

@article{takagi2023transformer,
  title={Transformer-based personalized attention mechanism for medical images with clinical records},
  author={Takagi, Yusuke and Hashimoto, Noriaki and Masuda, Hiroki and Miyoshi, Hiroaki and Ohshima, Koichi and Hontani, Hidekata and Takeuchi, Ichiro},
  journal={Journal of Pathology Informatics},
  volume={14},
  pages={100185},
  year={2023},
  publisher={Elsevier}
}

@article{hashimoto2024multimodal,
  title={Multimodal Gated Mixture of Experts Using Whole Slide Image and Flow Cytometry for Multiple Instance Learning Classification of Lymphoma},
  author={Hashimoto, Noriaki and Hanada, Hiroyuki and Miyoshi, Hiroaki and Nagaishi, Miharu and Sato, Kensaku and Hontani, Hidekata and Ohshima, Koichi and Takeuchi, Ichiro},
  journal={Journal of Pathology Informatics},
  volume={15},
  pages={100359},
  year={2024},
  publisher={Elsevier}
}

@article{taylor2018post,
  title={Post-selection inference for {$\ell_1$}-penalized likelihood models},
  author={Taylor, Jonathan and Tibshirani, Robert},
  journal={Canadian Journal of Statistics},
  volume={46},
  number={1},
  pages={41--61},
  year={2018},
  publisher={Wiley Online Library}
}

@article{ding2025multimodal,
  title={A multimodal whole-slide foundation model for pathology},
  author={Ding, Tong and Wagner, Sophia J and Song, Andrew H and Chen, Richard J and Lu, Ming Y and Zhang, Andrew and Vaidya, Anurag J and Jaume, Guillaume and Shaban, Muhammad and Kim, Ahrong and others},
  journal={Nature medicine},
  volume={31},
  number={11},
  pages={3749--3761},
  year={2025},
  publisher={Nature Publishing Group US New York}
}

@article{vorontsov2024foundation,
  title={A foundation model for clinical-grade computational pathology and rare cancers detection},
  author={Vorontsov, Eugene and Bozkurt, Alican and Casson, Adam and Shaikovski, George and Zelechowski, Michal and Severson, Kristen and Zimmermann, Eric and Hall, James and Tenenholtz, Neil and Fusi, Nicolo and others},
  journal={Nature medicine},
  volume={30},
  number={10},
  pages={2924--2935},
  year={2024},
  publisher={Nature Publishing Group US New York}
}

@article{fithian2014optimal,
  title={Optimal inference after model selection},
  author={Fithian, William and Sun, Dennis and Taylor, Jonathan},
  journal={arXiv preprint arXiv:1410.2597},
  year={2014}
}

@article{taylor2015statistical,
  title={Statistical learning and selective inference},
  author={Taylor, Jonathan and Tibshirani, Robert J},
  journal={Proceedings of the National Academy of Sciences},
  volume={112},
  number={25},
  pages={7629--7634},
  year={2015},
  publisher={National Academy of Sciences},
  doi={10.1073/pnas.1507583112}
}

@article{lee2016exact,
  title={Exact post-selection inference, with application to the lasso},
  author={Lee, Jason D and Sun, Dennis L and Sun, Yuekai and Taylor, Jonathan E},
  journal={The Annals of Statistics},
  volume={44},
  number={3},
  pages={907--927},
  year={2016},
  publisher={Institute of Mathematical Statistics},
  doi={10.1214/15-AOS1371}
}

@article{tibshirani2016exact,
  title={Exact post-selection inference for sequential regression procedures},
  author={Tibshirani, Ryan J and Taylor, Jonathan and Lockhart, Richard and Tibshirani, Robert},
  journal={Journal of the American Statistical Association},
  volume={111},
  number={514},
  pages={600--620},
  year={2016},
  publisher={Taylor \& Francis},
  doi={10.1080/01621459.2015.1108848}
}

@article{terada2023selective,
title={Selective inference after feature selection via multiscale bootstrap},
author={Terada, Yoshikazu and Shimodaira, Hidetoshi},
journal={Annals of the Institute of Statistical Mathematics},
volume={75},
number={1},
pages={99--125},
year={2023},
publisher={Springer}
}

@article{hyun2018exact,
title={Exact post-selection inference for the generalized lasso path},
author={Hyun, Sangwon and G'Sell, Max and Tibshirani, Ryan J},
journal={Electronic Journal of Statistics},
volume={12},
number={1},
pages={1053--1097},
year={2018},
publisher={Institute of Mathematical Statistics and Bernoulli Society}
}

@article{panigrahi2024selective,
title={Selective inference for sparse multitask regression with applications in neuroimaging},
author={Panigrahi, Snigdha and Stewart, Natasha and Sripada, Chandra and Levina, Elizaveta},
journal={The Annals of Applied Statistics},
volume={18},
number={1},
pages={445--467},
year={2024},
publisher={Institute of Mathematical Statistics}
}

@inproceedings{suzumura2017selective,
title={Selective inference for sparse high-order interaction models},
author={Suzumura, Shinya and Nakagawa, Kazuya and Umezu, Yuta and Tsuda, Koji and Takeuchi, Ichiro},
booktitle={International Conference on Machine Learning},
pages={3338--3347},
year={2017},
organization={PMLR}
}

@inproceedings{yamada2018post,
title={Post selection inference with kernels},
author={Yamada, Makoto and Umezu, Yuta and Fukumizu, Kenji and Takeuchi, Ichiro},
booktitle={International Conference on Artificial Intelligence and Statistics},
pages={152--160},
year={2018},
organization={PMLR}
}

@article{charkhi2018asymptotic,
title={Asymptotic post-selection inference for the Akaike information criterion},
author={Charkhi, Ali and Claeskens, Gerda},
journal={Biometrika},
volume={105},
number={3},
pages={645--664},
year={2018},
publisher={Oxford University Press}
}

@inproceedings{shiraishi2025statistical,
title={Statistical test for feature selection pipelines by selective inference},
author={Shiraishi, Tomohiro and Matsukawa, Tatsuya and Nishino, Shuichi and Takeuchi, Ichiro},
booktitle={International Conference on Machine Learning},
pages={55283--55302},
year={2025},
organization={PMLR}
}

@article{chen2023selective,
  title={Selective inference for k-means clustering},
  author={Chen, Yiqun T. and Witten, Daniela M.},
  journal={Journal of Machine Learning Research},
  volume={24},
  number={152},
  pages={1--41},
  year={2023}
}

@article{gao2024selective,
  title={Selective inference for hierarchical clustering},
  author={Gao, Lucy L. and Bien, Jacob and Witten, Daniela},
  journal={Journal of the American Statistical Association},
  volume={119},
  number={545},
  pages={332--342},
  year={2024},
  publisher={Taylor \& Francis},
  doi={10.1080/01621459.2022.2116331}
}

@inproceedings{duy2020computing,
  title={Computing Valid p-value for Optimal Changepoint by Selective Inference using Dynamic Programming},
  author={Duy, Vo Nguyen Le and Toda, Hiroki and Sugiyama, Ryota and Takeuchi, Ichiro},
  booktitle={Advances in Neural Information Processing Systems},
  volume={33},
  pages={11356--11367},
  year={2020},
  publisher={Curran Associates, Inc.}
}

@article{chen2020valid,
  title={Valid inference corrected for outlier removal},
  author={Chen, Shuxiao and Bien, Jacob},
  journal={Journal of Computational and Graphical Statistics},
  volume={29},
  number={2},
  pages={323--334},
  year={2020},
  publisher={Taylor \& Francis},
  doi={10.1080/10618600.2019.1660180}
}

@article{perry2026inference,
  title={Inference on the proportion of variance explained in principal component analysis},
  author={Perry, Ronan and Panigrahi, Snigdha and Bien, Jacob and Witten, Daniela},
  journal={Journal of the American Statistical Association},
  volume={121},
  number={553},
  pages={667--677},
  year={2026},
  publisher={Taylor \& Francis},
  doi={10.1080/01621459.2025.2538895}
}

@article{tian2018selective,
  title={Selective inference with a randomized response},
  author={Tian, Xiaoying and Taylor, Jonathan},
  journal={The Annals of Statistics},
  volume={46},
  number={2},
  pages={679--710},
  year={2018},
  publisher={Institute of Mathematical Statistics},
  doi={10.1214/17-AOS1564}
}

@inproceedings{le2021parametric,
  title={Parametric Programming Approach for More Powerful and General Lasso Selective Inference},
  author={Nguyen Le Duy, Vo and Takeuchi, Ichiro},
  booktitle={Proceedings of The 24th International Conference on Artificial Intelligence and Statistics},
  pages={901--909},
  year={2021},
  editor={Banerjee, Arindam and Fukumizu, Kenji},
  volume={130},
  series={Proceedings of Machine Learning Research},
  publisher={PMLR},
  url={https://proceedings.mlr.press/v130/nguyen-le-duy21a.html}
}

@inproceedings{sugiyama2021more,
  title={More Powerful and General Selective Inference for Stepwise Feature Selection using Homotopy Method},
  author={Sugiyama, Kazuya and Duy, Vo Nguyen Le and Takeuchi, Ichiro},
  booktitle={Proceedings of the 38th International Conference on Machine Learning},
  pages={9891--9901},
  year={2021},
  editor={Meila, Marina and Zhang, Tong},
  volume={139},
  series={Proceedings of Machine Learning Research},
  publisher={PMLR},
  url={https://proceedings.mlr.press/v139/sugiyama21a.html}
}

@article{le2022more,
  title={More Powerful Conditional Selective Inference for Generalized Lasso by Parametric Programming},
  author={Nguyen Le Duy, Vo and Takeuchi, Ichiro},
  journal={Journal of Machine Learning Research},
  volume={23},
  number={300},
  pages={1--37},
  year={2022}
}

@inproceedings{duy2022quantifying,
  title={Quantifying Statistical Significance of Neural Network-based Image Segmentation by Selective Inference},
  author={Duy, Vo Nguyen Le and Iwazaki, Shogo and Takeuchi, Ichiro},
  booktitle={Advances in Neural Information Processing Systems},
  volume={35},
  pages={31627--31639},
  year={2022},
  publisher={Curran Associates, Inc.},
  doi={10.52202/068431-2293}
}

@inproceedings{miwa2023salient,
  title={Valid P-Value for Deep Learning-Driven Salient Region},
  author={Miwa, Daiki and Duy, Vo Nguyen Le and Takeuchi, Ichiro},
  booktitle={The Eleventh International Conference on Learning Representations},
  year={2023},
  url={https://openreview.net/forum?id=qihMOPw4Sf_}
}

@inproceedings{niihori2025quantifying,
  title={Quantifying Statistical Significance of Deep Nearest Neighbor Anomaly Detection via Selective Inference},
  author={Niihori, Mizuki and Nishino, Shuichi and Katsuoka, Teruyuki and Shiraishi, Tomohiro and Taji, Kouichi and Takeuchi, Ichiro},
  booktitle={Advances in Neural Information Processing Systems},
  volume={38},
  pages={173174--173203},
  year={2025},
  publisher={Curran Associates, Inc.},
  doi={10.52202/085713-5762}
}

@article{nishino2025statistical,
  title={Statistical Test for Saliency Maps of Graph Neural Networks via Selective Inference},
  author={Nishino, Shuichi and Shiraishi, Tomohiro and Katsuoka, Teruyuki and Takeuchi, Ichiro},
  journal={Transactions on Machine Learning Research},
  issn={2835-8856},
  year={2025},
  url={https://openreview.net/forum?id=5NkXTCVa7F}
}

@inproceedings{shiraishi2024statistical,
  title={Statistical Test for Attention Maps in Vision Transformers},
  author={Shiraishi, Tomohiro and Miwa, Daiki and Katsuoka, Teruyuki and Duy, Vo Nguyen Le and Taji, Kouichi and Takeuchi, Ichiro},
  booktitle={Proceedings of the 41st International Conference on Machine Learning},
  pages={45079--45096},
  year={2024},
  volume={235},
  series={Proceedings of Machine Learning Research},
  publisher={PMLR},
  url={https://proceedings.mlr.press/v235/shiraishi24a.html}
}

@article{shiraishi2026statistical,
  title={Statistical Test for Attention in Transformers for Images and Time Series},
  author={Shiraishi, Tomohiro and Miwa, Daiki and Katsuoka, Teruyuki and Duy, Vo Nguyen Le and Nishino, Shuichi and Taji, Kouichi and Takeuchi, Ichiro},
  journal={Journal of Machine Learning Research},
  volume={27},
  number={119},
  pages={1--43},
  year={2026}
}

@article{katsuoka2025si4onnx,
  title={si4onnx: A Python package for Selective Inference in Deep Learning Models},
  author={Katsuoka, Teruyuki and Shiraishi, Tomohiro and Miwa, Daiki and Nishino, Shuichi and Takeuchi, Ichiro},
  journal={arXiv preprint arXiv:2501.17415},
  year={2025},
  doi={10.48550/arXiv.2501.17415}
}

\clearpage

\appendix

\section{Summary of notation}
\label{app:notation}

This appendix summarizes the principal notation used throughout the paper.
Bold lowercase and uppercase letters denote vectors and random vectors,
respectively, while calligraphic letters denote sets, selection maps, or
selection outcomes. For a positive integer $N$, $[N]:=\{1,\ldots,N\}$.

\subsection{Multiple instance learning}

\begin{center}
\begin{tabular}{p{0.27\linewidth}p{0.67\linewidth}}
\hline
Symbol & Description \\
\hline
$N$, $M^n$ & Number of bags and number of instances in the $n$-th bag. \\
$X^n=\{\bm x^{n,m}\}_{m\in[M^n]}$ & The $n$-th bag and its constituent input instances. \\
$Y^n$, $y^{n,m}$, $\hat Y^n$ & Bag label, unobserved instance label, and predicted bag label. \\
$f$, $f_{\mathrm{enc}}$, $f_{\mathrm{att}}$, $f_{\mathrm{clf}}$ & MIL model, feature extractor, attention network, and classifier. \\
$\bm h^{n,m}$, $\bm h_{\mathrm{bag}}^n$ & Feature representation of instance $m$ and attention-pooled representation of bag $n$. \\
$a^{n,m}$, $a'^{\,n,m}$, $\tilde a^{n,m}$ & Pre-sigmoid attention logit, post-sigmoid attention score, and normalized attention weight. \\
$\tau$, $\mathcal C(X)$ &
Attention threshold and the set of indices of high-attention
instances in bag $X$. \\
$d$, $d'$ & Input dimension and encoded feature dimension. \\
\hline
\end{tabular}
\end{center}

\subsection{Test and reference instances}

\begin{center}
\begin{tabular}{p{0.27\linewidth}p{0.67\linewidth}}
\hline
Symbol & Description \\
\hline
$X_{\mathrm{test}}=\{\bm x_{\mathrm{test}}^i\}_{i\in[M_{\mathrm{test}}]}$ &
Test bag containing $M_{\mathrm{test}}$ observed instances. \\
$X_{\mathrm{ref}}=\{\bm x_{\mathrm{ref}}^j\}_{j\in[M_{\mathrm{ref}}]}$ &
Reference set containing $M_{\mathrm{ref}}$ observed normal instances. \\
$\bm h_{\mathrm{test}}^i$, $\bm h_{\mathrm{ref}}^j$ &
Encoded features of test instance $i$ and reference instance $j$. \\
$a_{\mathrm{test}}^i$ &
Pre-sigmoid attention logit of test instance $i$. \\
$k$, $j_{\mathrm{med}}$ &
Number of nearest neighbors and index of the selected medoid reference instance. \\
$\bm X_{\mathrm{test}}$, $\bm X_{\mathrm{ref}}^j$ &
Random vector corresponding to the selected test instance and random vectors corresponding to the reference instances. \\
$\bm s_{\mathrm{test}}$, $\bm s_{\mathrm{ref}}^j$ &
Unknown signal vectors of the selected test instance and reference instance $j$. \\
$\bm\epsilon_{\mathrm{test}}$, $\bm\epsilon_{\mathrm{ref}}^j$ &
Independent Gaussian noise vectors with covariance $\sigma^2 I_d$. \\
$\sigma^2$, $I_d$ &
Noise variance and the $d$-dimensional identity matrix. \\
${\mathrm H}_0$, ${\mathrm H}_1$ &
Null and alternative hypotheses comparing the signal vectors of the selected test instance and selected medoid reference instance. \\
\hline
\end{tabular}
\end{center}

\subsection{Test statistics and selective inference}

\begin{center}
\begin{tabular}{p{0.27\linewidth}p{0.67\linewidth}}
\hline
Symbol & Description \\
\hline
$\bm y$, $\bm Y$ &
Concatenated observed vector and corresponding random vector formed from one test instance and all reference instances. \\
$\operatorname{vec}(\cdot)$ &
Vector concatenation operator used to construct $\bm y$ and $\bm Y$. \\
$P$ &
Orthogonal projection matrix representing the contrast between the test instance and the selected medoid reference instance. \\
$T(\bm Y)$, $\widetilde T(\bm Y)$ &
Test statistic $\|P\bm Y\|_2$ and its standardized version $\|P\bm Y\|_2/\sigma$. \\
$p_{\mathrm{naive}}$, $p_{\mathrm{selective}}$ &
Naive $p$-value and selective $p$-value. \\
$p_{\mathrm{Bonferroni}}$ &
Bonferroni-corrected $p$-value. \\
$\mathcal A(\bm Y)$, $\mathcal A(\bm y)$ &
Attention selection map and its observed outcome. \\
$\mathcal K(\bm Y)$, $\mathcal K(\bm y)$ &
$k$-nearest-neighbor retrieval map and its observed set. \\
$\mathcal M(\bm Y)$, $\mathcal M(\bm y)$ &
Adaptive reference-selection map, defined by $k$-nearest-neighbor retrieval followed by medoid selection, and its observed outcome. \\
$\mathcal E(\bm Y)$, $\mathcal E(\bm y)$ &
Overall selection map $(\mathcal A,\mathcal M)$ and its observed outcome. \\
$\mathcal Q(\bm Y)$, $\mathcal Q(\bm y)$ &
Nuisance statistic and its observed value, consisting of the direction of $P\bm Y$ and the component orthogonal to the range of $P$. \\
$\alpha$ &
Significance level. \\
$\lambda_{\mathrm{att}}$ &
Multiplicity factor accounting for attention-based target selection in the Bonferroni correction. \\
\hline
\end{tabular}
\end{center}

\subsection{One-dimensional conditional path and selection events}

\begin{center}
\begin{tabular}{p{0.27\linewidth}p{0.67\linewidth}}
\hline
Symbol & Description \\
\hline
$\bm a$, $\bm b$, $z$, $\bm Y(z)$ &
Quantities defining the conditional path
$\bm Y(z)=\bm a+\bm b z$, where
$z=\widetilde T(\bm Y)>0$. \\
$\bm X_{\mathrm{test}}(z)$, $\bm X_{\mathrm{ref}}^j(z)$ &
Conditional-path representations of the test instance and the $j$-th reference instance. \\
$j_{\mathrm{med}}^{\mathrm{obs}}$ &
Index of the observed representative reference,
$j_{\mathrm{med}}^{\mathrm{obs}}=\mathcal M(\bm y)$. \\
$\mathcal Z_{\mathcal A}$, $\mathcal Z_{\mathcal M}$ &
Feasible regions preserving the observed attention-selection outcome and selected representative reference, respectively. \\
$\mathcal Z_{\mathcal K}(K)$ &
Local region on the conditional path over which the
$k$-nearest-neighbor set is fixed to $K$. \\
$\mathcal Z_{\mathcal M\mid\mathcal K}(K)$ &
Local region within $\mathcal Z_{\mathcal K}(K)$ in which
$j_{\mathrm{med}}^{\mathrm{obs}}$ is selected as the medoid. \\
$\mathcal Z$ &
Overall truncation region
$\mathcal Z_{\mathcal A}\cap\mathcal Z_{\mathcal M}$. \\
$\bm\alpha_{\mathrm{test}}$, $\bm\beta_{\mathrm{test}}$ &
Affine coefficients of the encoded test feature on the conditional path. \\
$\bm\alpha_{\mathrm{ref}}^j$, $\bm\beta_{\mathrm{ref}}^j$ &
Affine coefficients of the encoded feature of reference instance $j$. \\
$D^j(z)$, $D_r^j$ $(r=0,1,2)$ &
Squared feature-space distance from the test instance to reference instance $j$ and its constant, linear, and quadratic coefficients. \\
$S_K^j(z)$, $S_{K,r}^j$ $(r=0,1,2)$ &
Medoid score of candidate $j$ within a fixed $k$-nearest-neighbor set $K$ and its constant, linear, and quadratic coefficients. \\
$\ell_r$, $u_r$ &
Lower and upper endpoints of the $r$-th interval constituting the truncation region $\mathcal Z$. \\
$R$ &
Number of disjoint intervals constituting the truncation region $\mathcal Z$. \\
\hline
\end{tabular}
\end{center}

\section{Proofs of theorems}

\paragraph{Noise variance.}
The theoretical guarantees assume that the true noise variance
$\sigma^2$ is known.
In the synthetic and MNIST-based experiments, the true variance used
for data generation is also used for inference.
For CAMELYON16, an estimate obtained from independent normal instances
is substituted for the true variance.
The proofs below do not account for uncertainty in this estimate and
do not establish exact finite-sample validity for this plug-in
implementation.

\subsection{Proof of Theorem~\ref{theorem:truncated}}
\label{app:proof_truncated}

Fix the observed selection outcome
$\mathcal E(\bm y)$ and let $P$ denote the corresponding deterministic
projection matrix throughout this proof.
Recall that, under the proposed conditioning scheme,
$\mathcal E=(\mathcal A,\mathcal M)$; the intermediate
$k$-nearest-neighbor set is not itself included in the conditioning
event.
The unconditional distributional calculations below use this fixed
matrix before conditioning on the selection event.

We next condition on
$\mathcal Q(\bm Y)=\mathcal Q(\bm y)$,
which gives
\begin{align*}
\mathcal Q(\bm Y)
=
\mathcal Q(\bm y)
\quad\Longleftrightarrow\quad
\left(
\frac{P\bm Y}{\|P\bm Y\|_2},
(I-P)\bm Y
\right)
=
\left(
\frac{P\bm y}{\|P\bm y\|_2},
(I-P)\bm y
\right)
\quad\Longleftrightarrow\quad
\bm Y
=
\bm a+\bm b z,
\end{align*}
where
\begin{align*}
\bm a
&=
(I-P)\bm y,
\\
\bm b
&=
\sigma
\frac{P\bm y}{\|P\bm y\|_2},
\\
z
&=
\widetilde T(\bm Y)
\in
\mathbb R_{>0}.
\end{align*}
Therefore,
\begin{align*}
&
\left\{
\bm Y
\in
\mathbb R^{(M_{\mathrm{ref}}+1)d}
~
\middle|
~
\mathcal E(\bm Y)
=
\mathcal E(\bm y),
~
\mathcal Q(\bm Y)
=
\mathcal Q(\bm y)
\right\}
\\
&=
\left\{
\bm a+\bm b z
~
\middle|
~
z\in\mathcal Z
\right\},
\end{align*}
where
\begin{equation}
\mathcal Z
=
\left\{
z\in\mathbb R_{>0}
~
\middle|
~
\mathcal E(\bm a+\bm b z)
=
\mathcal E(\bm y)
\right\}
\end{equation}
is the truncation region determined by the observed selection event.

Under the Gaussian model and the null hypothesis associated with the
fixed selection outcome,
\[
P\mathbb E[\bm Y]=\bm 0.
\]
Before conditioning on the selection event,
\[
P\bm Y
\sim
\mathcal N(\bm 0,\sigma^2 P).
\]
Since $P$ is an orthogonal projection matrix of rank $d$,
\[
\widetilde T(\bm Y)
=
\frac{\|P\bm Y\|_2}{\sigma}
\sim
\chi_d.
\]
Moreover, because $P$ and $I-P$ are orthogonal projections,
$P\bm Y$ and $(I-P)\bm Y$ are independent.
The norm of $P\bm Y$ is also independent of its direction
$P\bm Y/\|P\bm Y\|_2$.
Therefore,
$\widetilde T(\bm Y)$ is independent of
$\mathcal Q(\bm Y)$ before conditioning on the selection event.

Conditioning additionally on
$\mathcal E(\bm Y)=\mathcal E(\bm y)$
restricts the support of
$\widetilde T(\bm Y)$ to
$\mathcal Z$.
Consequently,
\begin{equation}
\widetilde T(\bm Y)
\mid
\mathcal E(\bm Y)=\mathcal E(\bm y),
~
\mathcal Q(\bm Y)=\mathcal Q(\bm y)
\end{equation}
follows a $\chi$ distribution with $d$ degrees of freedom truncated to
$\mathcal Z$.
\hfill$\square$

\subsection{Proof of Theorem~\ref{theorem:significance}}
\label{app:proof_selectivep}

By Theorem~\ref{theorem:truncated}, under the null hypothesis,
\begin{equation}
\widetilde T(\bm Y)
\mid
\mathcal E(\bm Y)=\mathcal E(\bm y),
~
\mathcal Q(\bm Y)=\mathcal Q(\bm y)
\end{equation}
follows a $\chi$ distribution with $d$ degrees of freedom truncated to
$\mathcal Z$.
Since the selective $p$-value is defined as the upper-tail probability
of this conditional distribution, the probability integral transform
gives
\begin{equation}
{\mathbb P}_{{\mathrm H}_0}
\left(
p_{\mathrm{selective}}
\le
\alpha
\;\middle|\;
\mathcal E(\bm Y)=\mathcal E(\bm y),
~
\mathcal Q(\bm Y)=\mathcal Q(\bm y)
\right)
=
\alpha,
\end{equation}
for any
$\alpha\in(0,1)$.

Since this equality holds for all possible values of
$\mathcal Q(\bm Y)$,
the law of iterated expectations gives
\begin{align*}
&
{\mathbb P}_{{\mathrm H}_0}
\left(
p_{\mathrm{selective}}
\le
\alpha
\;\middle|\;
\mathcal E(\bm Y)=\mathcal E(\bm y)
\right)
\\
&=
{\mathbb E}_{{\mathrm H}_0}
\left[
{\mathbb P}_{{\mathrm H}_0}
\left(
p_{\mathrm{selective}}
\le
\alpha
\;\middle|\;
\mathcal E(\bm Y)=\mathcal E(\bm y),
~
\mathcal Q(\bm Y)
\right)
\;\middle|\;
\mathcal E(\bm Y)=\mathcal E(\bm y)
\right]
\\
&=
\alpha.
\end{align*}
Therefore, the selective $p$-value exactly controls the conditional
Type-I error rate given the observed selection event.
This guarantee holds for each selection outcome for which the
corresponding null hypothesis is true.
\hfill$\square$

\section{Selection event characterization}
\label{app:selection_event}

In this appendix, we explicitly characterize the selection event
introduced in Section~\ref{sec:selection_event}.
Recall that the overall selection map is
\begin{equation}
\mathcal E(\bm Y)
=
\left(
\mathcal A(\bm Y),
\mathcal M(\bm Y)
\right),
\end{equation}
where $\mathcal A$ represents attention-based target selection and
$\mathcal M$ represents adaptive reference selection.
The reference-selection map $\mathcal M$ is determined by first
retrieving the $k$ nearest reference instances and then selecting
their medoid.
The intermediate $k$-nearest-neighbor set is not itself included in
the conditioning event.

According to Theorem~\ref{theorem:truncated}, after conditioning on
$\mathcal Q(\bm Y)=\mathcal Q(\bm y)$,
the random vector $\bm Y$ can be represented as
\begin{equation}
\bm Y(z)
=
\bm a+\bm b z,
\qquad
z\in\mathbb R_{>0},
\label{eq:appendix_affine_path}
\end{equation}
where
$z=\widetilde T(\bm Y)$
is the only remaining random quantity.
Let
$\bm X_{\mathrm{test}}(z)$
and
$\bm X_{\mathrm{ref}}^j(z)$
denote the test instance and the $j$-th reference instance contained
in $\bm Y(z)$, respectively.

Along the one-dimensional path in
Eq.~\eqref{eq:appendix_affine_path},
the observed selection outcome is preserved when the
attention-selection outcome and the final selected representative
reference remain unchanged.
The $k$-nearest-neighbor set may change along the path.
We therefore use regions with a fixed $k$-nearest-neighbor set only as
local partitions for characterizing the adaptive reference-selection
event.

\subsection{Attention selection event}
\label{app:attention_event}

The attention selection event determines whether the test instance
is selected according to its pre-sigmoid attention logit.
Along the one-dimensional path
$\bm Y(z)$,
the attention logit of the selected test instance is
\begin{equation}
f_{\mathrm{att}}
\left(
f_{\mathrm{enc}}
\left(
\bm X_{\mathrm{test}}(z)
\right)
\right).
\label{eq:attention_logit_path}
\end{equation}

The threshold $\tau$ is fixed before performing selective inference.
Depending on the experimental setting, it may either be specified
directly or determined from an independent dataset.
Once fixed, $\tau$ is treated as a constant throughout the selective
inference procedure.

The observed attention-selection outcome is preserved when
\begin{equation}
\mathbb I
\left\{
f_{\mathrm{att}}
\left(
f_{\mathrm{enc}}
\left(
\bm X_{\mathrm{test}}(z)
\right)
\right)
>
\tau
\right\}
=
\mathcal A(\bm y).
\label{eq:attention_selection_event}
\end{equation}

Since both
$f_{\mathrm{enc}}$
and
$f_{\mathrm{att}}$
are piecewise affine,
the attention logit is affine in
$z$
within each activation region.
Therefore, the feasible region associated with the attention
selection event is
\begin{equation}
\mathcal Z_{\mathcal A}
=
\left\{
z\in\mathbb R_{>0}
~
\middle|
~
\mathcal A(\bm a+\bm b z)
=
\mathcal A(\bm y)
\right\}.
\label{eq:attention_feasible_region}
\end{equation}

\subsection{$k$-nearest-neighbor partition}
\label{app:knn_event}

For each reference instance, define its squared Euclidean distance from
the test instance along the conditional path by
\begin{align}
D^j(z)
=
\left\|
f_{\mathrm{enc}}
\left(
\bm X_{\mathrm{test}}(z)
\right)
-
f_{\mathrm{enc}}
\left(
\bm X_{\mathrm{ref}}^j(z)
\right)
\right\|_2^2.
\label{eq:knn_distance}
\end{align}
Using squared Euclidean distances does not change the nearest-neighbor
ordering compared with ordinary Euclidean distances.

For any index set
$K\subseteq[M_{\mathrm{ref}}]$
with
$|K|=k$,
the $k$-nearest-neighbor set is equal to $K$ when every reference
instance in $K$ is no farther from the test instance than every
candidate outside $K$:
\begin{align}
D^j(z)
\le
D^{j'}(z),
\qquad
j\in K,
\quad
j'\in[M_{\mathrm{ref}}]\setminus K.
\label{eq:knn_ordering_condition}
\end{align}

Within each activation region, the encoded features are affine
functions of $z$:
\begin{align}
f_{\mathrm{enc}}
\left(
\bm X_{\mathrm{test}}(z)
\right)
&=
\bm\alpha_{\mathrm{test}}
+
\bm\beta_{\mathrm{test}}z,
\label{eq:test_feature_affine}
\\
f_{\mathrm{enc}}
\left(
\bm X_{\mathrm{ref}}^j(z)
\right)
&=
\bm\alpha_{\mathrm{ref}}^j
+
\bm\beta_{\mathrm{ref}}^jz.
\label{eq:reference_feature_affine}
\end{align}
Here,
$\bm\alpha_{\mathrm{test}}$,
$\bm\beta_{\mathrm{test}}$,
$\bm\alpha_{\mathrm{ref}}^j$, and
$\bm\beta_{\mathrm{ref}}^j$
are fixed within the corresponding activation region and are obtained
by propagating $\bm a$ and $\bm b$ through the encoder.

Substituting
Eqs.~\eqref{eq:test_feature_affine}--\eqref{eq:reference_feature_affine}
into Eq.~\eqref{eq:knn_distance} gives
\begin{align}
D^j(z)
&=
\left\|
\left(
\bm\alpha_{\mathrm{test}}
-
\bm\alpha_{\mathrm{ref}}^j
\right)
+
\left(
\bm\beta_{\mathrm{test}}
-
\bm\beta_{\mathrm{ref}}^j
\right)z
\right\|_2^2
\nonumber\\
&=
D_0^j
+
D_1^jz
+
D_2^jz^2,
\label{eq:knn_distance_quadratic}
\end{align}
where
\begin{align}
D_0^j
&=
\left\|
\bm\alpha_{\mathrm{test}}
-
\bm\alpha_{\mathrm{ref}}^j
\right\|_2^2,
\\
D_1^j
&=
2
\left(
\bm\alpha_{\mathrm{test}}
-
\bm\alpha_{\mathrm{ref}}^j
\right)^\top
\left(
\bm\beta_{\mathrm{test}}
-
\bm\beta_{\mathrm{ref}}^j
\right),
\\
D_2^j
&=
\left\|
\bm\beta_{\mathrm{test}}
-
\bm\beta_{\mathrm{ref}}^j
\right\|_2^2.
\end{align}

Therefore, each ordering condition in
Eq.~\eqref{eq:knn_ordering_condition}
is equivalent to the quadratic inequality
\begin{align}
\left(
D_0^j-D_0^{j'}
\right)
+
\left(
D_1^j-D_1^{j'}
\right)z
+
\left(
D_2^j-D_2^{j'}
\right)z^2
\le
0.
\label{eq:knn_quadratic_constraint}
\end{align}

For a fixed $k$-nearest-neighbor set $K$, define the corresponding
local region by
\begin{align}
\mathcal Z_{\mathcal K}(K)
=
\bigcap_{\substack{
j\in K\\
j'\in[M_{\mathrm{ref}}]\setminus K
}}
\left\{
z\in\mathbb R_{>0}
\;\middle|\;
D^j(z)-D^{j'}(z)\le0
\right\}.
\label{eq:knn_feasible_region}
\end{align}
The regions
$\mathcal Z_{\mathcal K}(K)$
partition the conditional path according to the
$k$-nearest-neighbor set, up to boundary points determined by the
prespecified tie-breaking rule.
Importantly,
$\mathcal Z_{\mathcal K}(K)$
is used only as a local partition for characterizing the
reference-selection map;
the $k$-nearest-neighbor set itself is not conditioned upon.

If $k$-nearest-neighbor retrieval is performed in the input space,
the same derivation applies by replacing the encoded test and
reference features with
$\bm X_{\mathrm{test}}(z)$ and
$\bm X_{\mathrm{ref}}^j(z)$,
respectively.

\subsection{Medoid selection event}
\label{app:medoid_event}

Within a region where the $k$-nearest-neighbor set is fixed to $K$,
the medoid is the reference instance that minimizes the sum of squared
Euclidean distances to the other instances in $K$.

For each candidate $j\in K$, define its medoid score by
\begin{align}
S_K^j(z)
=
\sum_{j'\in K}
\left\|
f_{\mathrm{enc}}
\left(
\bm X_{\mathrm{ref}}^j(z)
\right)
-
f_{\mathrm{enc}}
\left(
\bm X_{\mathrm{ref}}^{j'}(z)
\right)
\right\|_2^2.
\label{eq:medoid_score}
\end{align}

Let
\begin{equation}
j_{\mathrm{med}}^{\mathrm{obs}}
=
\mathcal M(\bm y)
\end{equation}
denote the observed representative reference.
For a fixed $K$ containing
$j_{\mathrm{med}}^{\mathrm{obs}}$,
the observed reference-selection outcome is preserved when
\begin{align}
S_K^{j_{\mathrm{med}}^{\mathrm{obs}}}(z)
\le
S_K^j(z),
\qquad
j\in
K\setminus
\left\{
j_{\mathrm{med}}^{\mathrm{obs}}
\right\}.
\label{eq:medoid_ordering_condition}
\end{align}
If
$j_{\mathrm{med}}^{\mathrm{obs}}\notin K$,
the observed representative-reference outcome cannot be preserved
within that $k$-nearest-neighbor region.

Using the affine representation in
Eq.~\eqref{eq:reference_feature_affine},
each medoid score can be written as
\begin{align}
S_K^j(z)
&=
\sum_{j'\in K}
\left\|
\left(
\bm\alpha_{\mathrm{ref}}^j
-
\bm\alpha_{\mathrm{ref}}^{j'}
\right)
+
\left(
\bm\beta_{\mathrm{ref}}^j
-
\bm\beta_{\mathrm{ref}}^{j'}
\right)z
\right\|_2^2
\nonumber\\
&=
S_{K,0}^j
+
S_{K,1}^j z
+
S_{K,2}^j z^2,
\label{eq:medoid_score_quadratic}
\end{align}
where
\begin{align}
S_{K,0}^j
&=
\sum_{j'\in K}
\left\|
\bm\alpha_{\mathrm{ref}}^j
-
\bm\alpha_{\mathrm{ref}}^{j'}
\right\|_2^2,
\\
S_{K,1}^j
&=
2
\sum_{j'\in K}
\left(
\bm\alpha_{\mathrm{ref}}^j
-
\bm\alpha_{\mathrm{ref}}^{j'}
\right)^\top
\left(
\bm\beta_{\mathrm{ref}}^j
-
\bm\beta_{\mathrm{ref}}^{j'}
\right),
\\
S_{K,2}^j
&=
\sum_{j'\in K}
\left\|
\bm\beta_{\mathrm{ref}}^j
-
\bm\beta_{\mathrm{ref}}^{j'}
\right\|_2^2.
\end{align}

Thus,
Eq.~\eqref{eq:medoid_ordering_condition}
reduces to
\begin{align}
\left(
S_{K,0}^{j_{\mathrm{med}}^{\mathrm{obs}}}
-
S_{K,0}^{j}
\right)
+
\left(
S_{K,1}^{j_{\mathrm{med}}^{\mathrm{obs}}}
-
S_{K,1}^{j}
\right)z
+
\left(
S_{K,2}^{j_{\mathrm{med}}^{\mathrm{obs}}}
-
S_{K,2}^{j}
\right)z^2
\le
0.
\label{eq:medoid_quadratic_constraint}
\end{align}

For a fixed $K$ containing
$j_{\mathrm{med}}^{\mathrm{obs}}$,
define the corresponding local medoid region by
\begin{align}
\mathcal Z_{\mathcal M\mid\mathcal K}(K)
=
\bigcap_{
j\in
K\setminus
\{j_{\mathrm{med}}^{\mathrm{obs}}\}
}
\left\{
z\in\mathbb R_{>0}
~
\middle|
~
S_K^{j_{\mathrm{med}}^{\mathrm{obs}}}(z)
-
S_K^j(z)
\le
0
\right\}.
\label{eq:local_medoid_feasible_region}
\end{align}

Since the $k$-nearest-neighbor set is not conditioned upon, the
overall region preserving the observed representative reference is
obtained by combining all $k$-nearest-neighbor regions for which the
same medoid is selected:
\begin{align}
\mathcal Z_{\mathcal M}
=
\bigcup_{\substack{
K\subseteq[M_{\mathrm{ref}}],\,|K|=k\\
j_{\mathrm{med}}^{\mathrm{obs}}\in K
}}
\left[
\mathcal Z_{\mathcal K}(K)
\cap
\mathcal Z_{\mathcal M\mid\mathcal K}(K)
\right].
\label{eq:medoid_feasible_region}
\end{align}
Thus, the $k$-nearest-neighbor set may change across different parts
of $\mathcal Z_{\mathcal M}$ as long as the final selected medoid
remains
$j_{\mathrm{med}}^{\mathrm{obs}}$.

If medoid selection is performed in the input space, the same
derivation applies by replacing the encoded reference features
$f_{\mathrm{enc}}(\bm X_{\mathrm{ref}}^j(z))$
with
$\bm X_{\mathrm{ref}}^j(z)$.

Throughout this appendix, an intersection over an empty index set is
interpreted as $\mathbb R_{>0}$.
In particular, this convention applies to
$\mathcal Z_{\mathcal K}(K)$ when $k=M_{\mathrm{ref}}$ and to
$\mathcal Z_{\mathcal M\mid\mathcal K}(K)$ when $k=1$.

\subsection{Overall selection event}
\label{app:overall_selection_event}

The overall selection outcome is represented by
\begin{equation}
\mathcal E(\bm Y)
=
\left(
\mathcal A(\bm Y),
\mathcal M(\bm Y)
\right).
\label{eq:appendix_overall_selection_map}
\end{equation}
Therefore, preserving the observed selection outcome
$\mathcal E(\bm y)$
requires simultaneous preservation of the attention-selection outcome
and the selected representative reference.
The $k$-nearest-neighbor set itself need not remain equal to its
observed value.

The resulting truncation region is
\begin{align}
\mathcal Z
&=
\left\{
z\in\mathbb R_{>0}
~
\middle|
~
\mathcal E(\bm a+\bm b z)
=
\mathcal E(\bm y)
\right\}
\nonumber\\
&=
\mathcal Z_{\mathcal A}
\cap
\mathcal Z_{\mathcal M}.
\label{eq:overall_truncation_region}
\end{align}

Within each activation region,
the attention constraint is characterized by linear inequalities,
whereas the boundaries associated with $k$-nearest-neighbor retrieval
and medoid selection are characterized by quadratic inequalities.
Because these inequalities may produce disconnected feasible subsets,
the overall truncation region generally takes the form
\begin{equation}
\mathcal Z
=
\bigcup_{r=1}^{R}
[\ell_r,u_r],
\label{eq:truncation_union}
\end{equation}
where
$R\in\mathbb N$
is the number of disjoint feasible intervals,
and
$\ell_r$
and
$u_r$
denote the lower and upper endpoints of the
$r$-th interval, respectively.
Whether individual endpoints are included does not affect the
selective $p$-value because the null distribution is continuous.

\subsection{Computation of the truncation region}
\label{app:truncation_computation}

The truncation region is computed along the one-dimensional path
$\bm Y(z)=\bm a+\bm b z$.
Within each activation region of the encoder and attention network,
the affine representations of the encoded features and attention
logits are fixed.
Consequently, the attention constraint and the boundaries associated
with $k$-nearest-neighbor retrieval and medoid selection reduce to
linear or quadratic inequalities in $z$.

The $k$-nearest-neighbor boundaries partition the conditional path
into local regions over which
$\mathcal K(\bm Y(z))$
is fixed.
Within each such region, the medoid constraints determine whether the
selected representative reference is equal to
$j_{\mathrm{med}}^{\mathrm{obs}}$.
Parametric programming explores these local regions along the
one-dimensional path and retains those portions satisfying the
observed attention-selection outcome and
\[
\mathcal M(\bm Y(z))
=
\mathcal M(\bm y).
\]
The retained intervals are then combined over all local regions and
merged when adjacent.
Importantly, a change in the $k$-nearest-neighbor set does not by
itself exclude an interval from the truncation region if the final
selected medoid remains unchanged.
In contrast to over-conditioning, which uses only the local feasible
interval containing the observed statistic, this procedure identifies
the entire truncation region consistent with the observed selection
outcome.

Given the decomposition in
Eq.~\eqref{eq:truncation_union},
the selective $p$-value is computed as
\begin{equation}
p_{\mathrm{selective}}
=
\frac{
\displaystyle
\sum_{
r:\,
u_r\ge\widetilde T(\bm y)
}
\int_{
\max\left\{
\ell_r,
\widetilde T(\bm y)
\right\}
}^{u_r}
f_{\chi_d}(z)\,dz
}{
\displaystyle
\sum_{r=1}^{R}
\int_{\ell_r}^{u_r}
f_{\chi_d}(z)\,dz
},
\label{eq:selective_p_union}
\end{equation}
where
$f_{\chi_d}$
denotes the probability density function of the
$\chi$ distribution with $d$ degrees of freedom.

Equation~\eqref{eq:selective_p_union}
is the upper-tail probability of the
$\chi_d$ distribution truncated to the selection-dependent region
$\mathcal Z$.

\section{Experimental details and additional results}
\label{app:exp_add}

This appendix provides additional experimental details that were omitted
from the main paper due to space limitations, together with additional
experimental results under various parameter settings.
For each dataset, we first describe the corresponding experimental
settings and then present supplementary results that complement the
findings reported in Section~5.

\paragraph{Computational environment.}
All experiments, except for UNI2 feature extraction, were implemented
in Python and executed on an x86\_64 Linux computer equipped with two
Intel Xeon Gold 6230 CPUs (20 cores each, 40 CPU cores in total).
UNI2 feature extraction for CAMELYON16 was performed on a separate
Linux computer equipped with an NVIDIA RTX A6000 GPU.

\subsection{Synthetic experiments}
\label{app:exp_synth}

\paragraph{Experimental settings.}
This section provides additional details of the synthetic data
experiments.
The inference settings are identical to those described in
Section~\ref{sec:exp_synth}.

For each combination of the input dimension $d$ and Gaussian noise
variance $\sigma^2$, we independently trained an ABMIL model using
synthetically generated bags.
Each training epoch consisted of 1,000 positive bags and 1,000 negative
bags, each containing 10 instances.
The number of positive instances in a positive bag was sampled uniformly
from one to ten, while negative bags consisted entirely of negative
instances.
New training bags were generated at every epoch.

The ABMIL model consisted of an MLP-based feature extractor, an
attention network, and a bag-level classifier.
The feature extractor had hidden dimensions $d/2$ and $d/4$, while both
the attention network and the classifier used a hidden dimension of
$d/8$.
ReLU activations were used in all hidden layers.
Attention weights were obtained by applying a sigmoid function followed
by a softmax normalization across the instances in a bag.

The model was trained for 10 epochs using the binary cross-entropy loss
and the Adam optimizer with an initial learning rate of $10^{-3}$.
The batch size was set to one, and the learning rate was halved every
five epochs.
After training, only the feature extractor and the attention network
were used in the subsequent selective inference experiments.

\paragraph{Additional results.}
Besides the experiments reported in the main paper, we additionally
evaluated the effects of the input dimension and the Gaussian noise
variance under different settings of the number of nearest neighbors
$k$.

\begin{figure}[tb]
\begin{center}
 \includegraphics[width=0.4\linewidth]
 {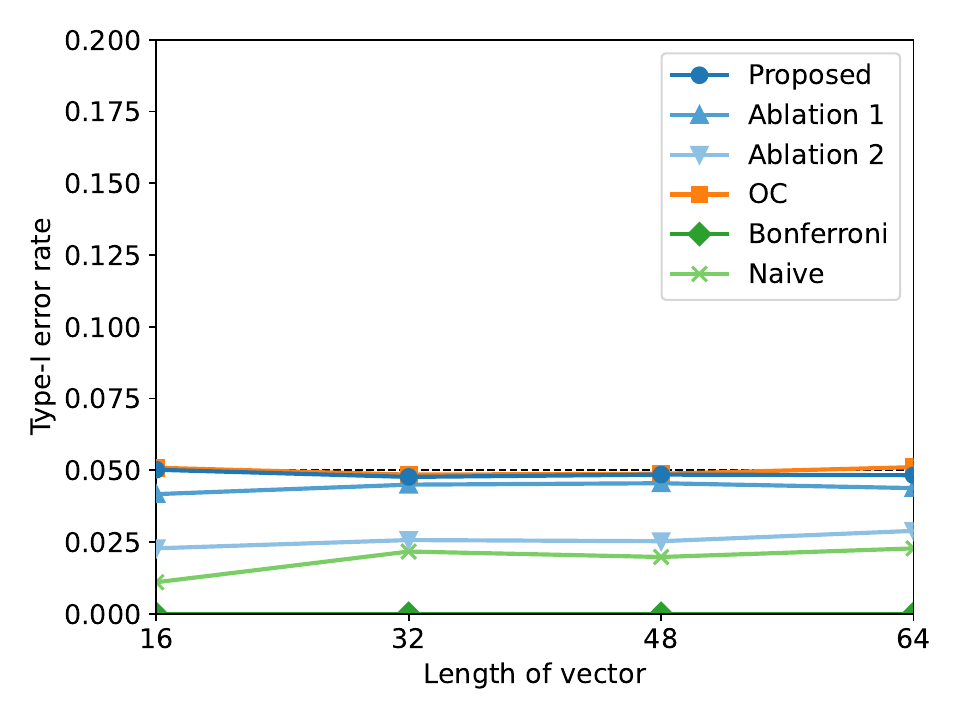}
 ~~~
 \includegraphics[width=0.4\linewidth]
 {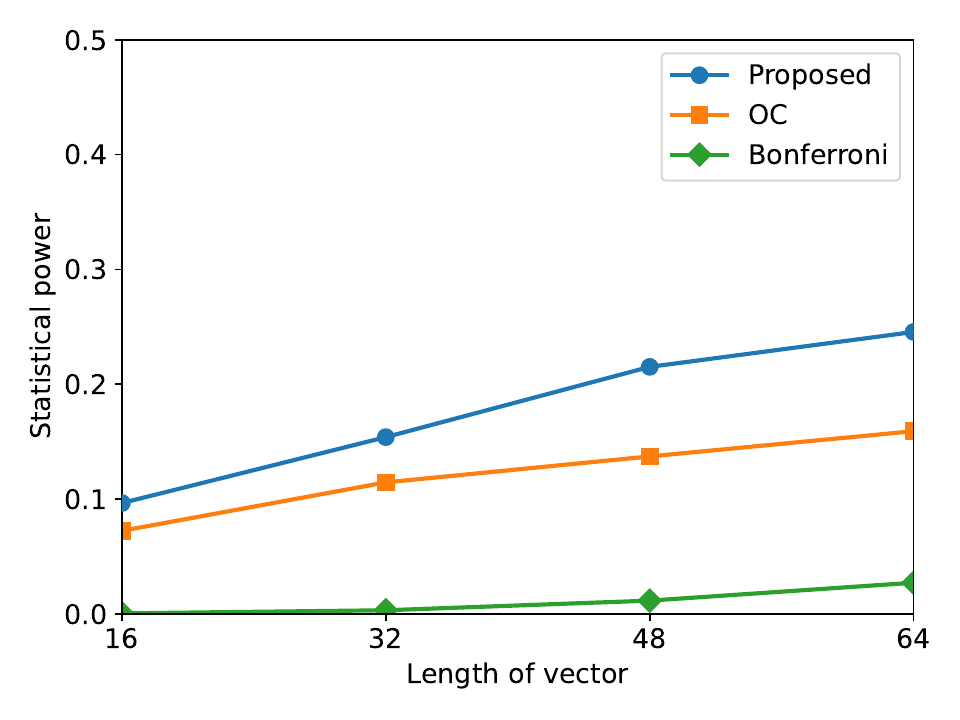}
 \\
 \includegraphics[width=0.4\linewidth]
 {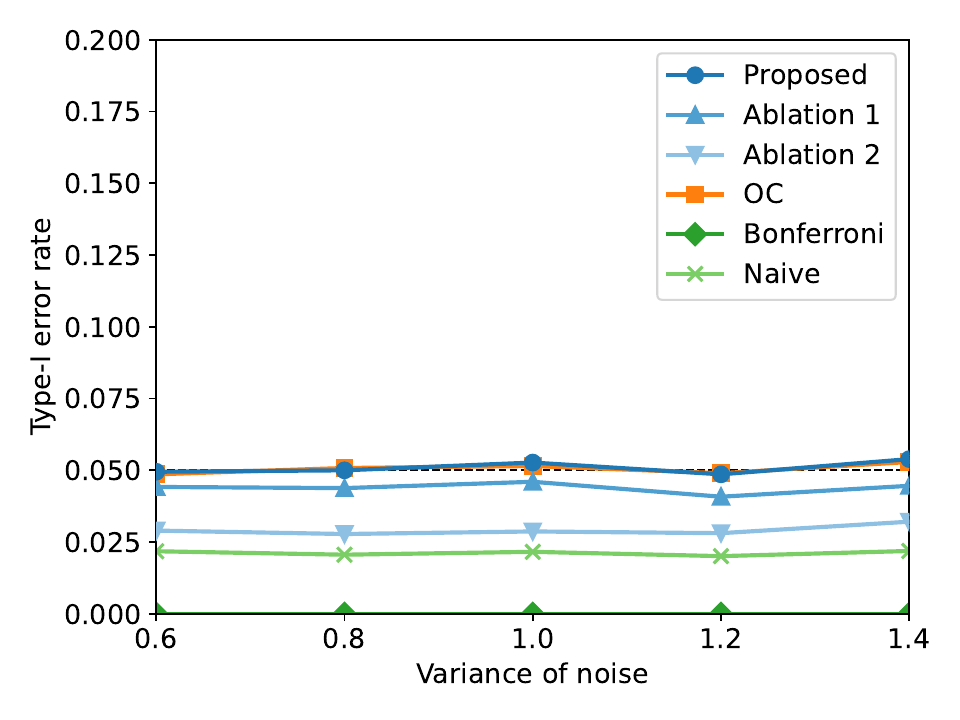}
 ~~~
 \includegraphics[width=0.4\linewidth]
 {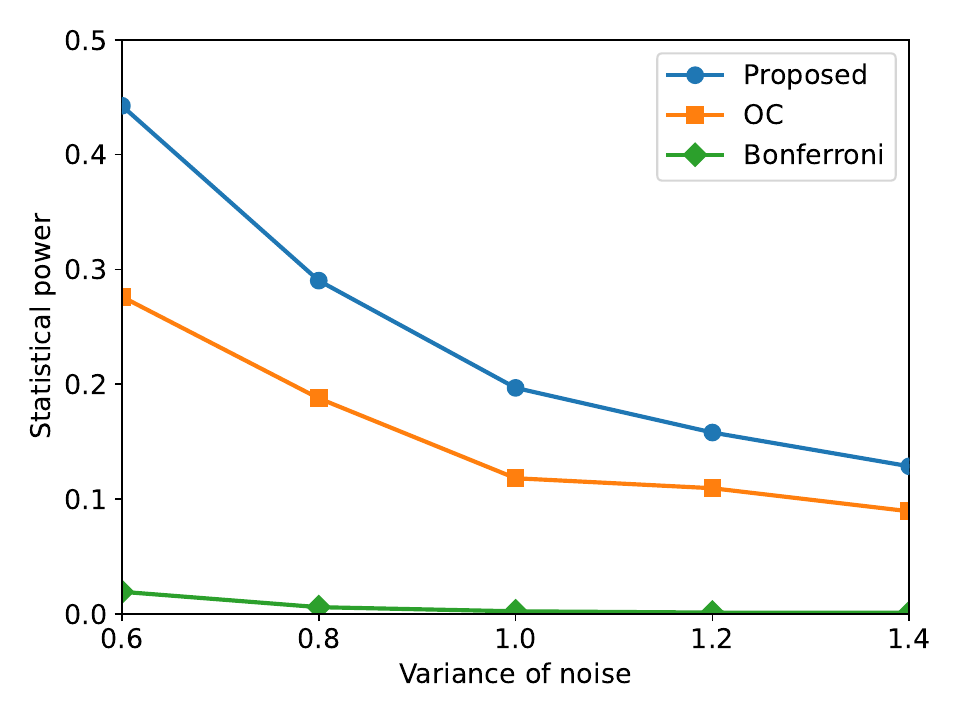}
  \\
 \includegraphics[width=0.4\linewidth]
 {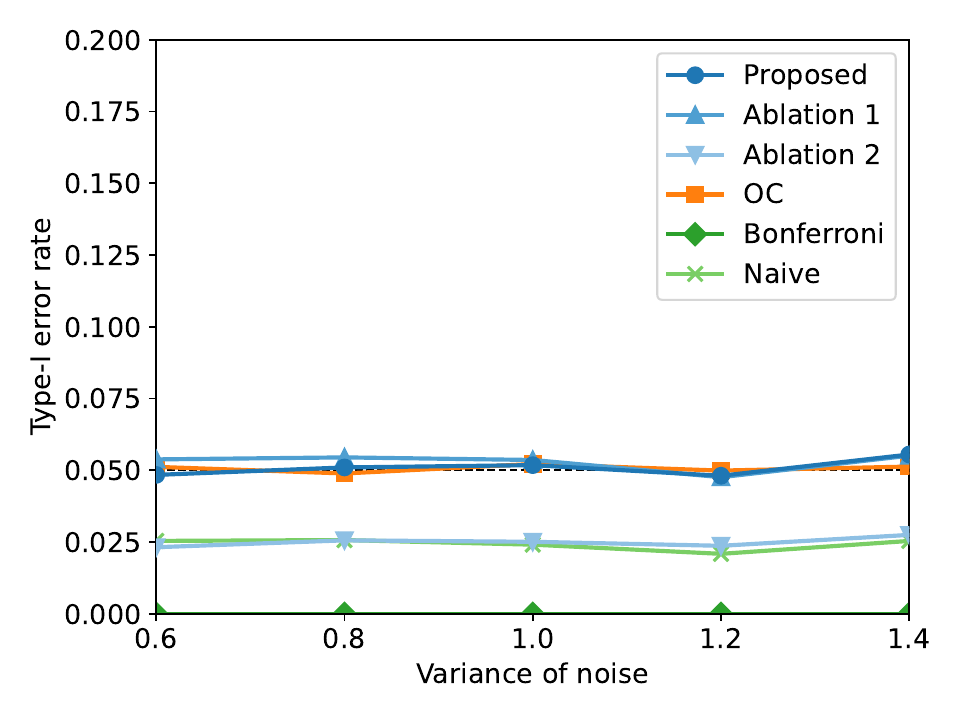}
 ~~~
 \includegraphics[width=0.4\linewidth]
 {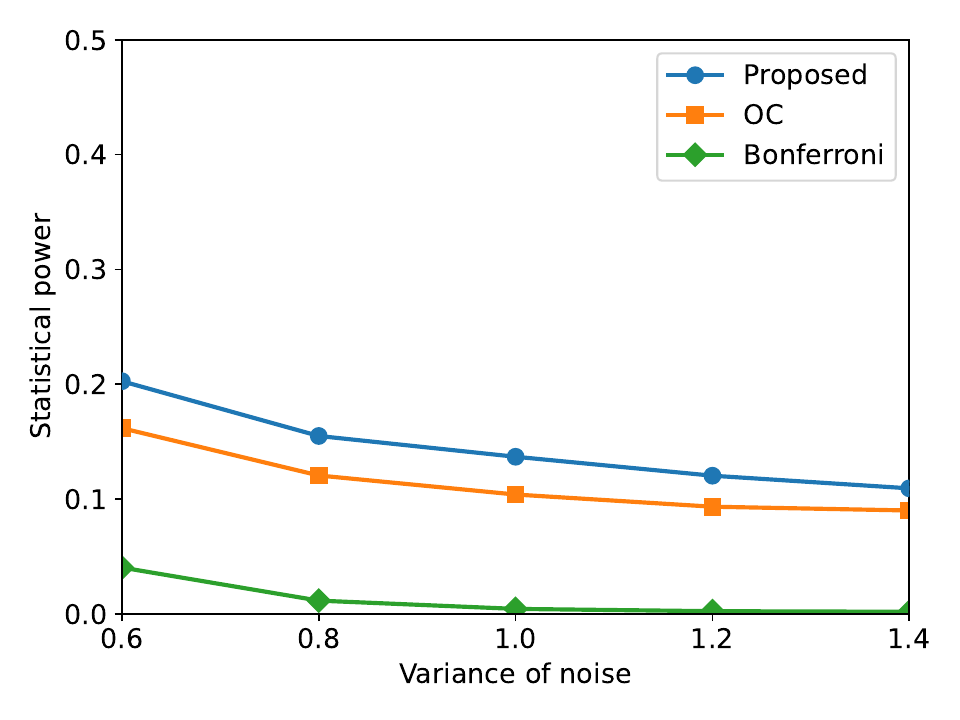}
\end{center}
\caption{
Type-I error rate (left column) and statistical power (right column)
under different parameter settings in the synthetic experiments.
The dashed horizontal line indicates the nominal significance level of
0.05.
The top row shows the results obtained by varying the input dimension
while fixing the Gaussian noise variance to $\sigma^2=1.0$ and the
number of nearest neighbors to $k=5$.
The middle row shows the results obtained by varying the Gaussian noise
variance while fixing the input dimension to 32 and setting $k=1$.
The bottom row shows the corresponding results with $k=10$.
}
\label{fig:exp_synth_add}
\end{figure}

The top row of Figure~\ref{fig:exp_synth_add} shows the results obtained
by varying the input dimension while fixing the Gaussian noise variance
to $\sigma^2=1.0$ and the number of nearest neighbors to $k=5$.
The proposed method consistently controlled the Type-I error rate close
to the nominal significance level.
Ablation~1 also exhibited Type-I error rates close to the nominal level
in the examined settings, whereas Ablation~2 showed conservative
behavior.
OC maintained valid Type-I error control, while Bonferroni correction
was highly conservative.
For the statistical power, the proposed method consistently
outperformed OC, and the difference between the two methods became
larger as the input dimension increased.
The statistical power of both methods increased with the input
dimension.

The middle and bottom rows show the results obtained by varying the
Gaussian noise variance while fixing the input dimension to $d=32$.
The middle row corresponds to $k=1$, whereas the bottom row corresponds
to $k=10$.
Across both settings, the proposed method maintained Type-I error rates
close to the nominal significance level.
Ablation~1 showed Type-I error rates around the nominal level in the
examined range of noise variances, whereas Ablation~2 remained
conservative.
OC also maintained valid Type-I error control, while Bonferroni
correction was highly conservative.
For the statistical power, the proposed method consistently
outperformed OC, and the statistical power decreased as the noise
variance increased for both methods.
Overall, these additional results confirm that the proposed method
maintains valid Type-I error control across a range of input dimensions,
noise variances, and values of $k$, while consistently achieving higher
statistical power than OC.

\FloatBarrier
\subsection{MNIST-based experiments}
\label{app:exp_mnist}

\paragraph{Experimental settings.}
This section provides additional details of the MNIST experiments.
The settings for statistical inference are identical to those described
in Section~\ref{sec:exp_mnist}.

For each experiment, digit ``0'' was used as the negative class,
while one of the digits $\{1,\ldots,9\}$ was specified as the
positive class.
After downsampling each image to $14\times14$ as described in
Section~\ref{sec:exp_mnist}, the images were flattened into
196-dimensional vectors and normalized to the range $[0,1]$.
For each class, a specified number of MNIST images were selected as
cluster centers.
An instance was generated by randomly selecting one of these centers
and adding independent Gaussian noise with variance $\sigma^2$ to each
dimension.
Separate cluster centers were selected from the MNIST training and test
sets for model training and statistical inference, respectively.

Each training epoch consisted of 1,000 positive bags and 1,000 negative
bags, with 10 instances in each bag.
The number of positive-class instances in a positive bag was sampled
uniformly from one to ten, and the remaining instances were generated
from the negative class.
A negative bag consisted entirely of negative-class instances.
New Gaussian noise was generated at every epoch.

The ABMIL model consisted of a linear feature extractor that mapped the
196-dimensional input to a 32-dimensional feature vector, followed by
an attention network and a bag-level classifier.
Both the attention network and classifier used a 16-dimensional hidden
layer with a ReLU activation.
As in the synthetic data experiments, the attention logits were
transformed by a sigmoid function and subsequently normalized across
the bag using softmax.

The model was trained for 10 epochs using the binary cross-entropy loss
and the Adam optimizer with an initial learning rate of $10^{-3}$.
The batch size was set to one, and the learning rate was halved every
five epochs.
After training, only the feature extractor and attention network were
used in the subsequent selective inference experiments.

\paragraph{Additional results.}
In addition to the results reported in the main paper, we evaluated
different values of the number of nearest neighbors $k$ under both the
three-cluster and five-cluster settings.
The positive and negative classes and the Gaussian noise variance were
kept unchanged from the corresponding experiments in
Section~\ref{sec:exp_mnist}.

\begin{figure}[tb]
\begin{center}
 \includegraphics[width=0.95\linewidth]{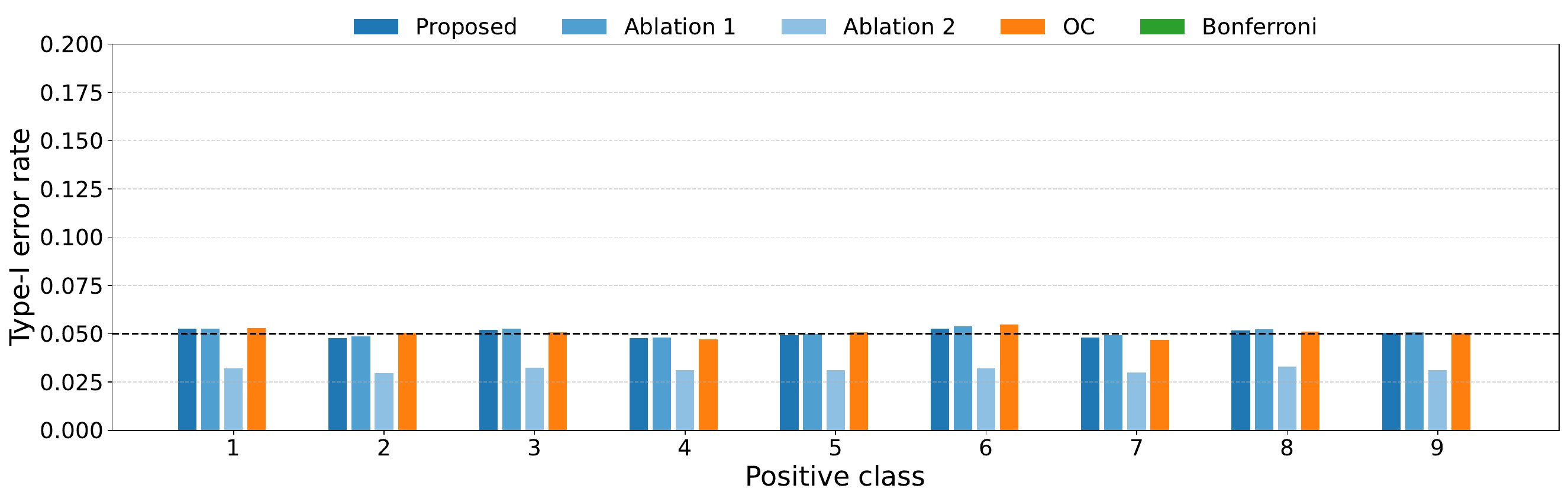}\\
 \includegraphics[width=0.95\linewidth]{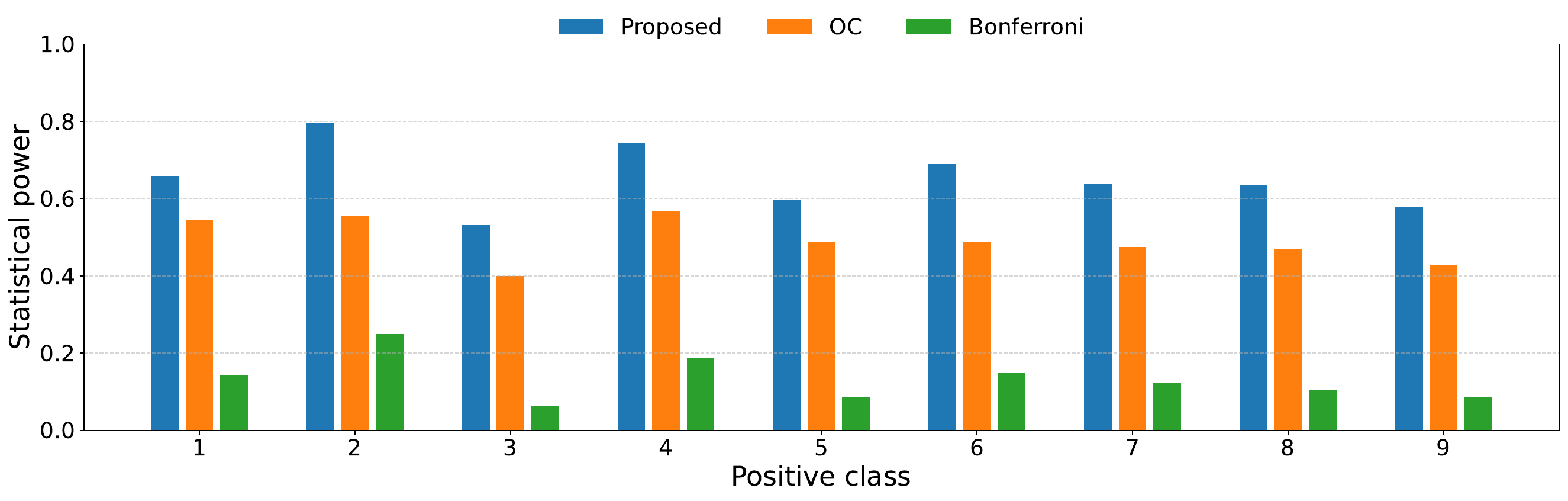}\\
 \includegraphics[width=0.95\linewidth]{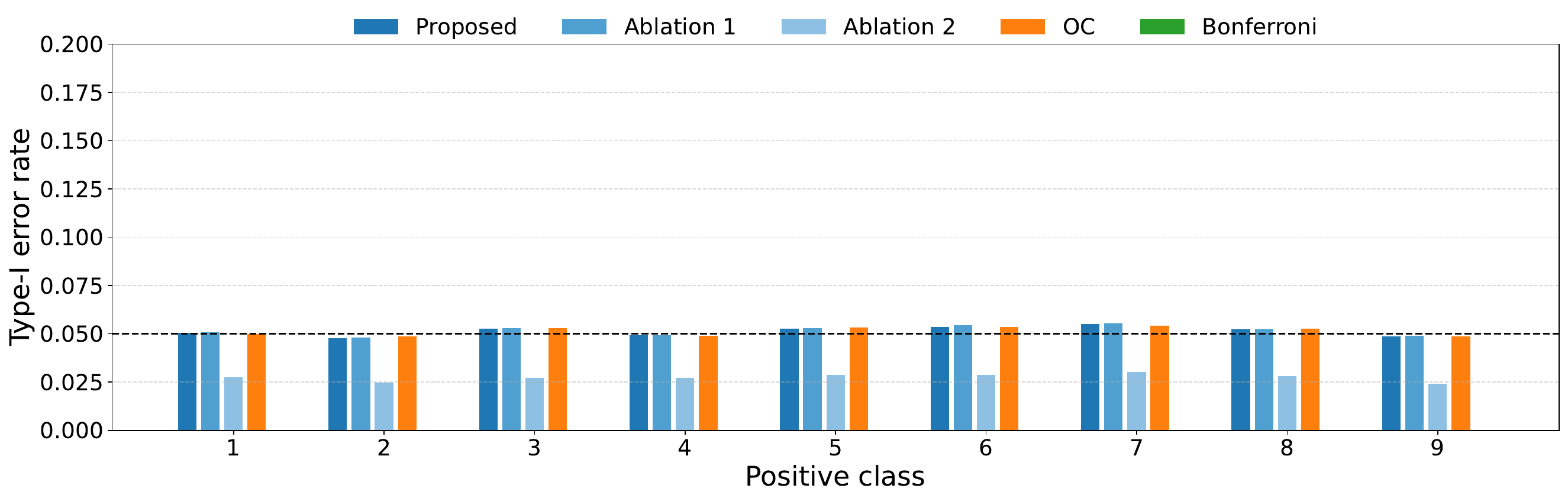}\\
 \includegraphics[width=0.95\linewidth]{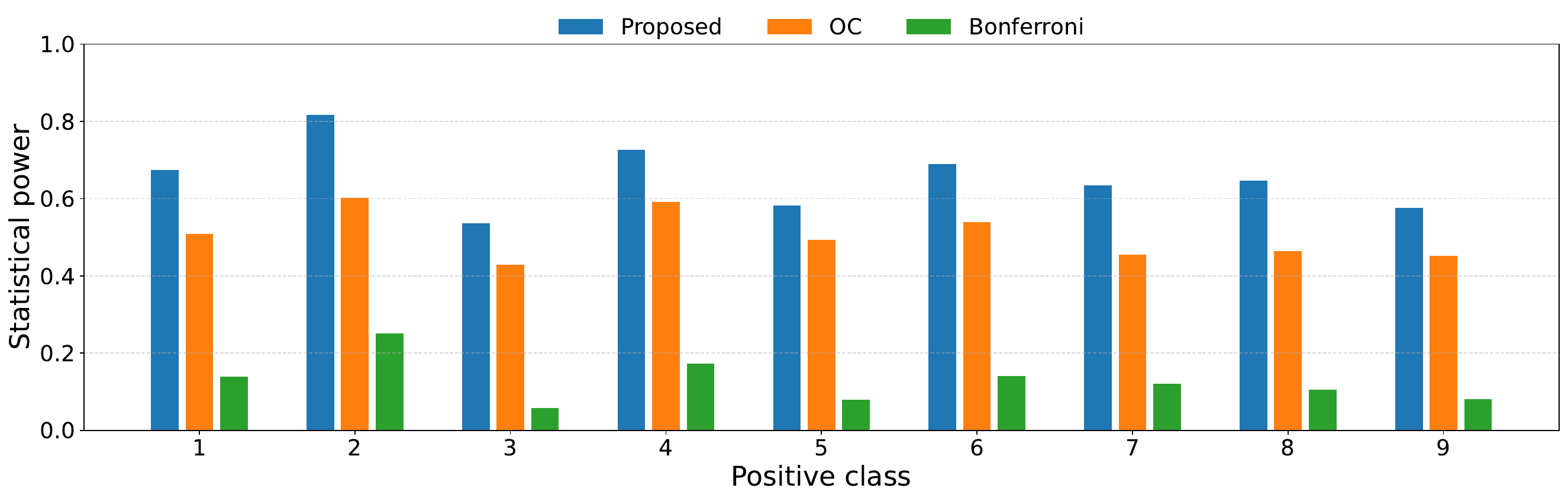}
\end{center}
 \caption{
Type-I error rate and statistical power in the MNIST-based experiments
using three Gaussian clusters.
From top to bottom, the plots show the Type-I error rate and statistical power with $k=3$, followed by the Type-I error rate and statistical power with $k=1$.
The horizontal axis represents the positive class.
}
\label{fig:exp_mnist_three}
\end{figure}

\begin{figure}[tb]
\begin{center}
 \includegraphics[width=0.95\linewidth]{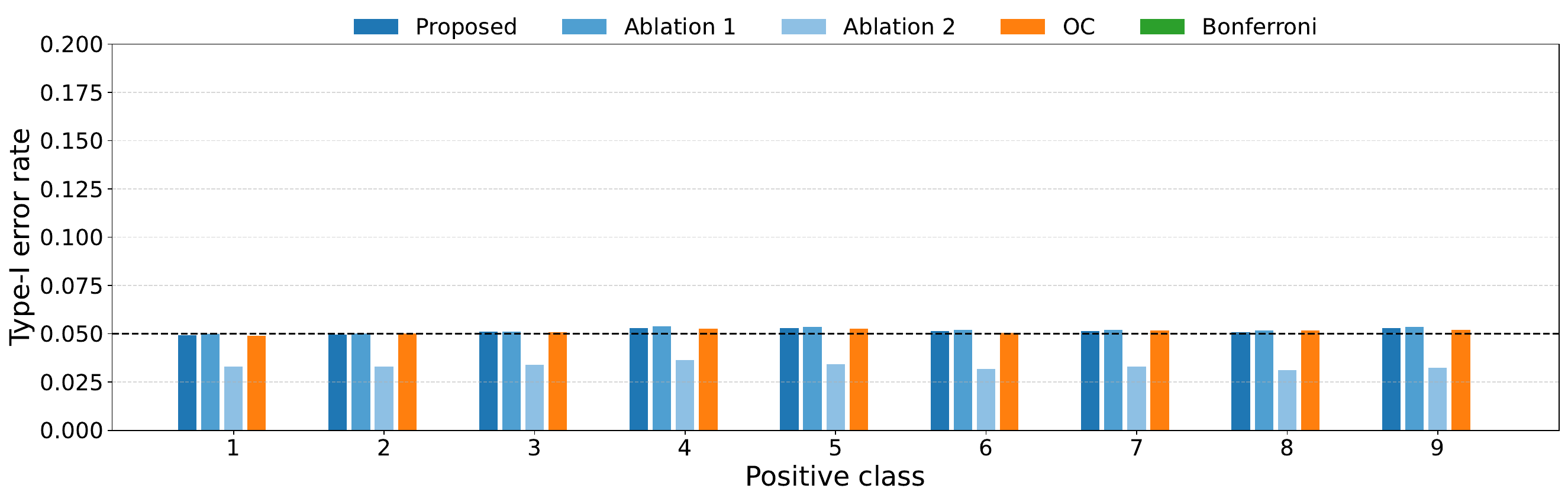}\\
 \includegraphics[width=0.95\linewidth]{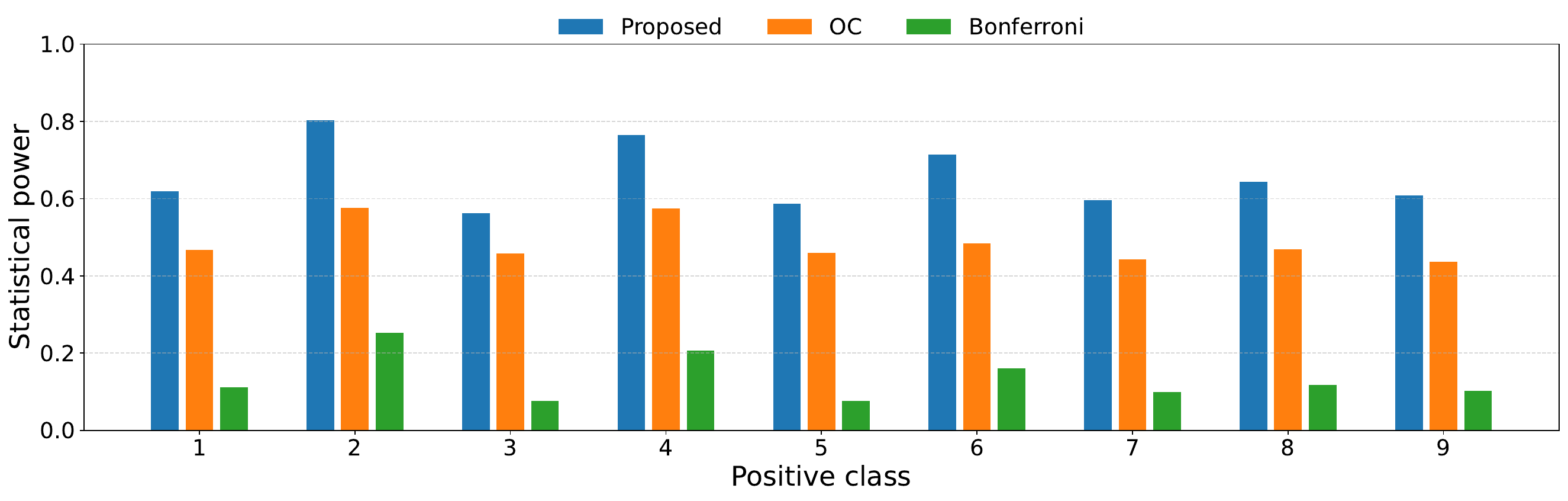}\\
 \includegraphics[width=0.95\linewidth]{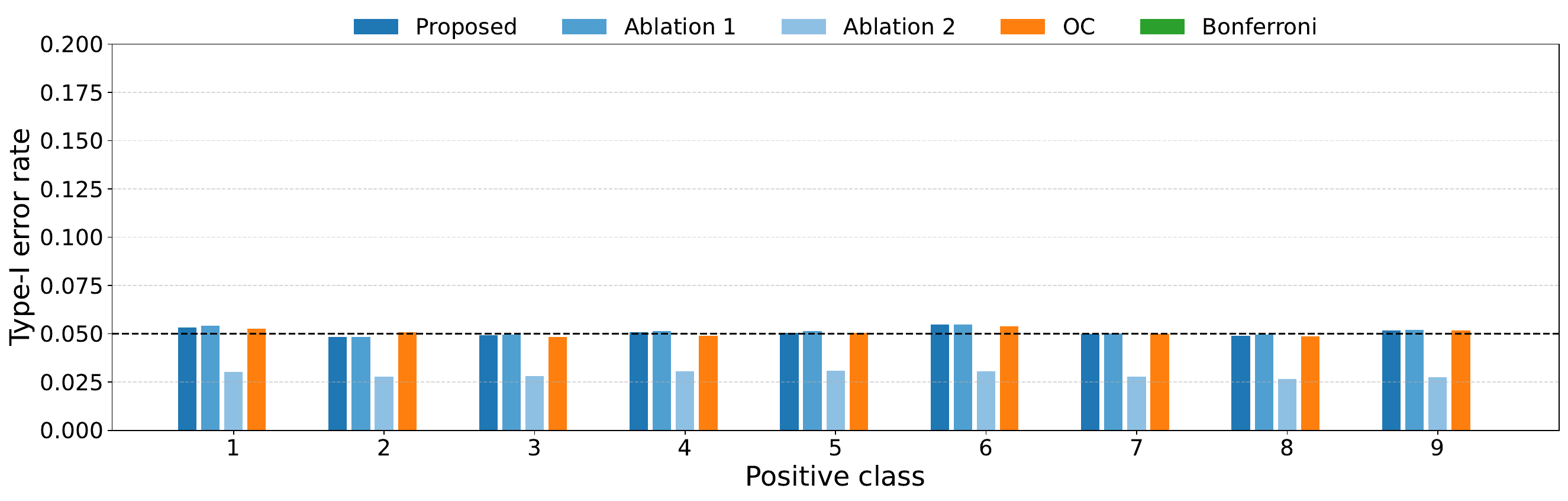}\\
 \includegraphics[width=0.95\linewidth]{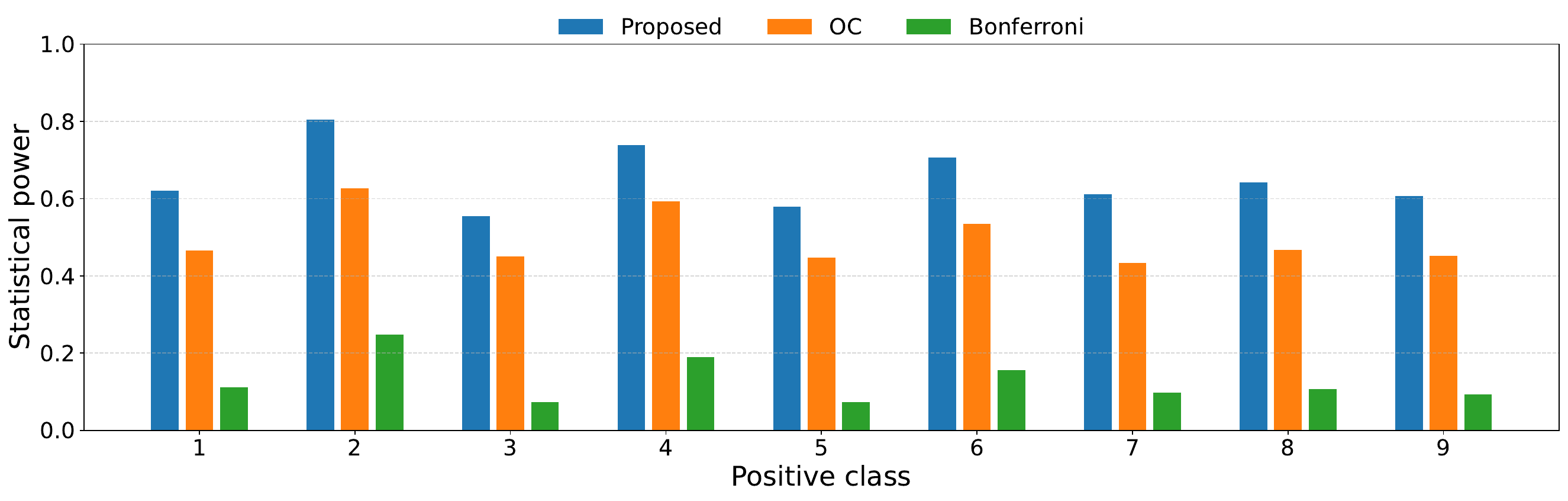}
\end{center}
 \caption{
Type-I error rate and statistical power in the MNIST-based experiments
using five Gaussian clusters.
From top to bottom, the plots show the Type-I error rate and statistical power with $k=3$, followed by the Type-I error rate and statistical power with $k=1$.
The horizontal axis represents the positive class.
}
\label{fig:exp_mnist_five}
\end{figure}

Figure~\ref{fig:exp_mnist_three} shows the Type-I error rate and
statistical power for the three-cluster setting, and
Figure~\ref{fig:exp_mnist_five} shows the corresponding results for the
five-cluster setting.
Across the examined values of $k$, the proposed method consistently
controlled the Type-I error rate close to the nominal significance
level while achieving higher statistical power than OC.
Bonferroni remained substantially conservative and markedly less
powerful than the proposed method and OC.
These results indicate that the performance of the proposed method is
robust to both the number of clusters and the choice of $k$.

\FloatBarrier
\subsection{Real-world experiments on CAMELYON16}
\label{app:exp_camelyon}

\paragraph{Experimental settings.}
This section provides additional implementation details of the
CAMELYON16 experiments.
The dataset construction, feature extraction,
reference construction, and selective inference settings are identical to those described in
Section~\ref{sec:exp_camelyon}.

The attention-based MIL model consisted of a linear feature extractor
that mapped the 192-dimensional pooled UNI2 features to a
32-dimensional representation, followed by an attention network and a
bag-level classifier with a 16-dimensional hidden layer.
ReLU activations were used in the hidden layers.
As in the synthetic and MNIST experiments, the attention logits were
transformed by a sigmoid function and normalized using softmax across
instances within each bag.

For model training, each WSI was randomly divided into bags consisting
of 50 image patches.
Up to 50 bags were generated from each WSI, depending on the number of
available patches, and the bag assignment was regenerated at every
training epoch.

The model was trained for 10 epochs using the binary cross-entropy loss
and the Adam optimizer with an initial learning rate of $10^{-4}$.
The batch size was set to one, and the learning rate was halved every
five epochs.
For slide-level evaluation, all patches from each WSI were treated as
a single bag.
The trained ABMIL model achieved a slide-level classification accuracy
of 0.897 on the test set.
After training, only the feature extractor and attention network were
used in the subsequent selective inference experiments.

\begin{figure*}[tb]
\begin{center}
\includegraphics[width=0.97\linewidth]{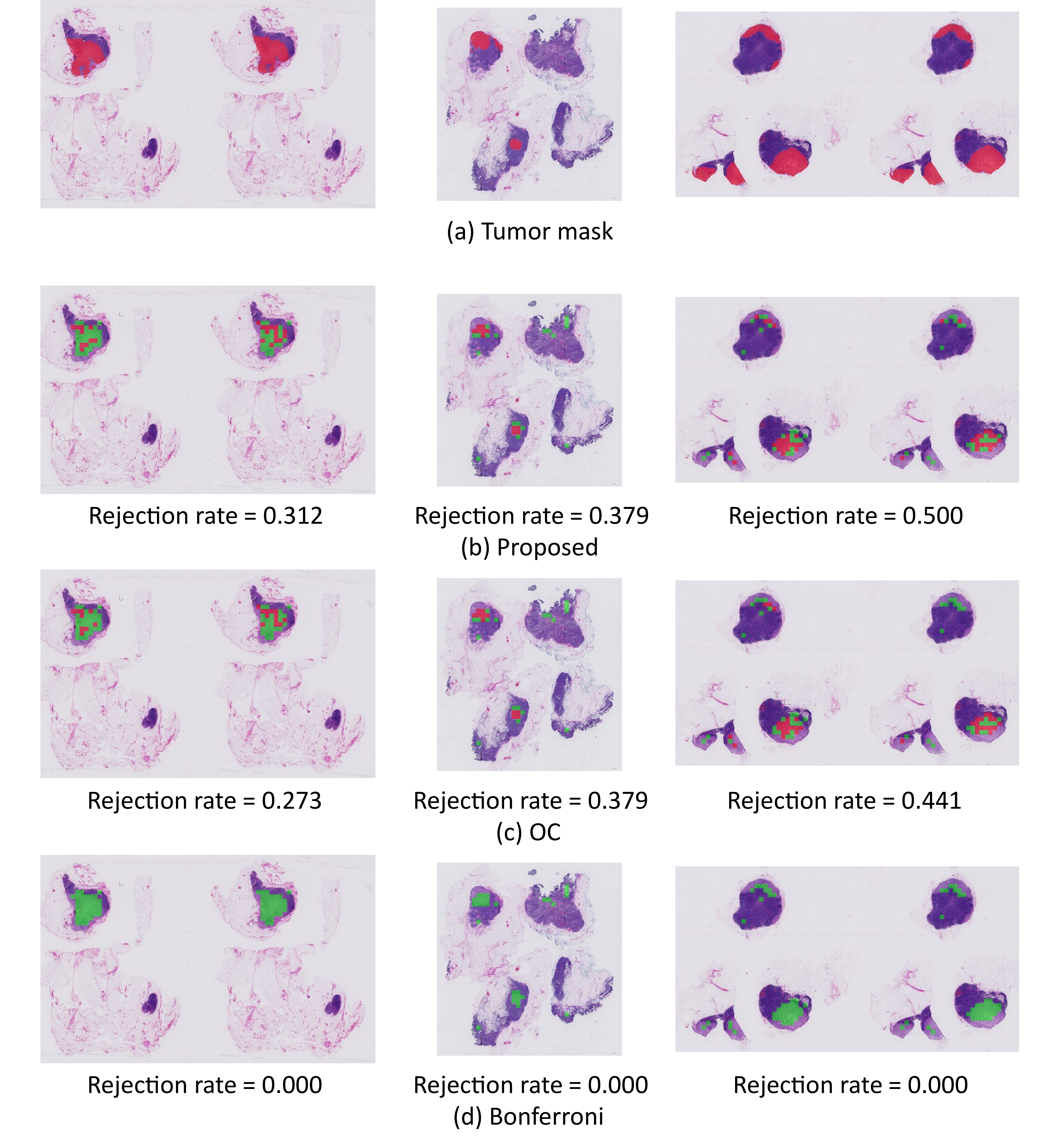}
\end{center}
 \caption{
Visualization of $p$-value maps for additional CAMELYON16 tumor WSIs
(Part I).
(a) shows the ground-truth tumor masks, while (b), (c), and (d) show
the results of Proposed, OC, and Bonferroni, respectively.
Green and red rectangles indicate selected high-attention instances
with $p \geq 0.05$ and $p < 0.05$, respectively.
}
\label{fig:exp_cam_add1}
\end{figure*}

\begin{figure*}[tb]
\begin{center}
\includegraphics[width=0.97\linewidth]{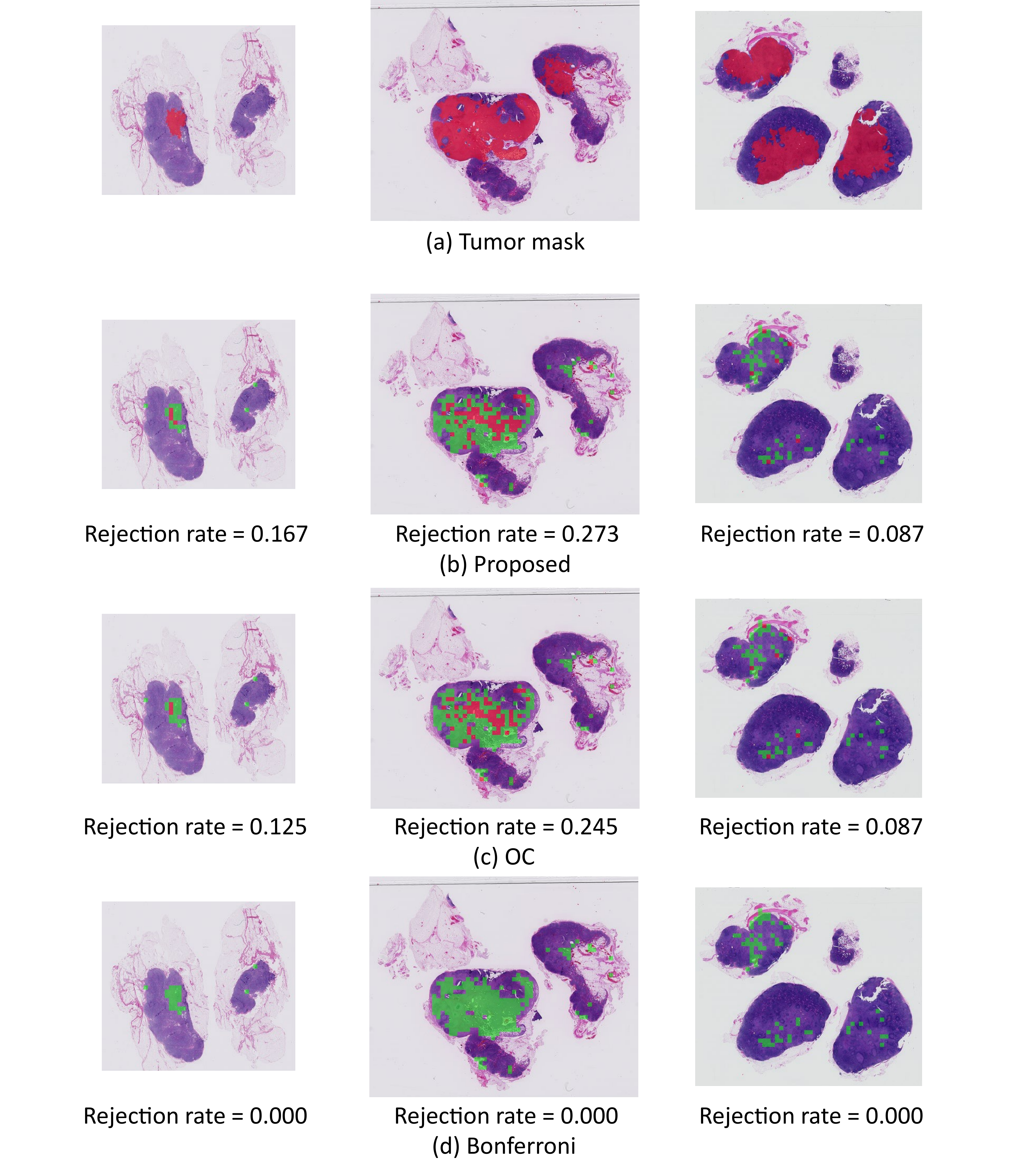}
\end{center}
 \caption{
Visualization of $p$-value maps for additional CAMELYON16 tumor WSIs
(Part II).
(a) shows the ground-truth tumor masks, while (b), (c), and (d) show
the results of Proposed, OC, and Bonferroni, respectively.
Green and red rectangles indicate selected high-attention instances
with $p \geq 0.05$ and $p < 0.05$, respectively.
}
\label{fig:exp_cam_add2}
\end{figure*}

\paragraph{Additional results.}
Additional qualitative results are presented in
Figures~\ref{fig:exp_cam_add1} and
\ref{fig:exp_cam_add2}.
The main paper presents the visualization results for three tumor
WSIs.
For completeness, we present the remaining six tumor WSIs in
the test set.
The proposed method generally identifies more statistically
significant regions than OC, while Bonferroni yields substantially fewer rejections.

\end{document}